\documentclass[conference]{IEEEtran}
\IEEEoverridecommandlockouts
\usepackage[utf8]{inputenc}
\usepackage[T1]{fontenc}
\usepackage{lmodern}
\usepackage{amsmath}
\usepackage{amssymb}
\usepackage{amsfonts}
\usepackage{amsthm}
\usepackage{graphicx}
\usepackage{booktabs}
\usepackage{array}
\usepackage{multirow}
\usepackage{xcolor}
\usepackage{cite}
\usepackage{url}
\usepackage[hidelinks]{hyperref}

\newtheorem{proposition}{Proposition}

\newcommand{\bus}{\text{bus}}
\newcommand{\inj}{\text{inj}}

\newcommand{\figorplaceholder}[2]{\IfFileExists{#2}{\includegraphics[width=\linewidth]{#2}}{\fbox{\parbox[c][3.2cm][c]{0.9\linewidth}{\centering\small\textbf{TODO:} #1}}}}

\begin{document}

\title{IP-GNO: An Inductive Physics-Informed Graph Neural Operator for Automated Distribution System State Estimation
\thanks{The authors are with the AMRIT Research Laboratory, Department of Electrical Engineering, Indian Institute of Technology (BHU), Varanasi, 221005, UP, India ((sakshi.kumar.cd.eee23, chhavi.jhari.eee23, ajaysingh.rs.eee24, subho.eee)@iitbhu.ac.in).}
\thanks{The research work is supported by the project titled ``Data-driven Advanced Management for Situational awareness in DER-rich partially Observable distribution Networks (DAMSON)” (File number: ANRF/ECRG/2024/000987/ENS), sponsored by Anusandhan National Research Foundation, India (Project ID: R\&D/ANRF/4211/EE/25-26/680).}
}

\author{\IEEEauthorblockN{Sakshi Kumar, Chhavi Jhari, Ajay Singh, \textit{SIEEE}, and Subho Paul, \textit{MIEEE}}
\vspace{-8mm}
}

\maketitle

\begin{abstract}
Distribution System State Estimation (DSSE) is a crucial tool for real-time monitoring, operation, and control of the new-age distribution systems (DSs) due to intensified voltage fluctuations and operational uncertainties introduced by volatile power injection at the nodes. 
Existing GNN-based approaches generally exhibit limited generalization across networks with different sizes and topologies, making zero-shot cross-grid transfer to unseen DSs a significant challenge. To overcome the aforementioned challenges, this article proposes an Inductive Physics-Informed Graph Neural Operator (IP-GNO) for scalable and transferable DSSE. The proposed framework frames complex bus voltage estimation as an operator-learning problem inspired by the Neumann series expansion of the inverse bus admittance matrix. An admittance-parameterized kernel, incorporating a self-admittance gate in the zeroth-order term and a persistent source-injection skip connection across network layers, captures the underlying electrical dependencies. Additionally, a Physics-Embedded Tree Encoding (PETE) scheme introduces five grid-size-invariant positional features derived from path impedance and a DC-linearized phase-angle prior. A sine-cosine output formulation is further employed to address phase-angle periodicity and ensure stable training. The proposed IP-GNO module is evaluated on two small (IEEE 33- and 69-bus), two medium (IEEE 123- and 136-bus), and one large (2522-bus) sized DSs. Particularly, its zero-shot cross-grid generalization capability is investigated by training the model exclusively on the IEEE 33-bus system and directly evaluating it on four unseen distribution networks of different sizes without retraining or fine-tuning. A detailed comparative analysis with the existing state-of-the-art DSSE tools proves the superiority of IP-GNO, enabling scalable DSSE across heterogeneous DSs. 
\end{abstract}

\begin{IEEEkeywords}
graph neural operator, physics-informed machine learning, distribution grid, power-flow surrogate, zero-shot generalization, bus injection model, Neumann series, operator learning
\end{IEEEkeywords}

\section{Introduction}

\subsection{Background and Motivations}
In the smart grid era, electrical distribution systems are undergoing a paradigm shift due to the rapid integration of volatile distributed energy resources and the growing demand due to the electrification of new customers and the adoption of additional electrical appliances by existing customers. As per the statistics published by the International Energy Agency (IEA), the worldwide electrification rate will rise to 35\% by 2035 \cite{IEA2026Electrification}. This increasing electricity demand causes expansion of the existing distribution system (DS) to ensure reliable power supply. Real-time monitoring and management of such large DSs becomes increasingly pivotal for achieving optimal control and efficient operation of the network \cite{9537302_Paul}.

In this regard, development of effective distribution system state estimation (DSSE) tools is often proposed as a crucial requirement for accurate and reliable estimation of the DS's operating states (specifically nodal voltage magnitude and angles) using available measurements from different sensory devices \cite{7779155_Primadianto}. 

\subsection{Prior Arts and Existing Research Challenges}
The state-of-the-art DSSE tools are classified into two major sections, viz. optimization-based and learning-based strategies. 

\subsubsection{\textbf{Optimization-Based DSSE Tools}}
Conventionally, DSSE tools are developed as a non-convex weighted least square (WLS) error optimization problem and solved iteratively using Gauss-Newton algorithm to minimize the gap between true and estimated value by representing the state variables in polar \cite{336098_Baran}, rectangular \cite{4538565_Nuqui}, and complex \cite{8264797_Dzafic} domains. Branch current based DSSE solvers are proposed in \cite{373974_Baran_Kelley, 1266570_Habin} to reduce the computation complexity in the presence of wye-connected load with solidly grounded neutral. To improve the performance of the WLS, Al-Wakeel et al. \cite{AlWakeel2016} proposed an iterative reweighted WLS (IRWLS), where unlike WLS, the measurement weights are modified at each iteration for obtaining fast convergence. Yilmaz et al. \cite{10173547_Yilmaz} developed parallel distributed SE process using least absolute and network partitioning (both radial and meshed) algorithms to solve the optimization problem iteratively for determining the state variables. A decentralized WLS concept is suggested in \cite{10900416_Alsyoufi} to perform local SE at the edge-computing devices. Bajagain and Dubey came up with maximum likelihood optimization method in \cite{11408920_Bajagain} to include skewed bimodal measurement error into the DSSE problem. Nanchian et al. \cite{7898803_Nanchian} employed the Ordinal optimization algorithm to determine the system states by minimizing the gap between the measured quantities and their functional relationship with the estimated states. 

The above methods are designed to solve a complex optimization problem integrated with non-convex bus injection model to relate current with voltage. To relax the computation complexities related to non-convex optimization, Soltani et al. \cite{9773346_Soltani} initially proposed a mixed-integer non-linear programming (MINLP)-based WLS SE problem considering the polar power-voltage (PPV) formulation and rectangular current-voltage (RIV) formulation. Later the authors revised the problem into mixed-integer quadratic programming (MIQP) by relaxing the non-convex relations using first order Taylor series expansion and big-M methods, and solved iteratively. Khan et al. \cite{9432710_Khan} proposed a linear SE algorithm by replacing the non-convex power flow expressions with their first order Taylor series expansion. Xu et al. \cite{11370943_Xu} developed an edge-enabled state estimation, where non-linear iterative estimation is solved with optimal initialization during abrupt measurement changing hours, and non-iterative linearized SE problem is solved during no-change hours. To avoid iterative solution process, Zargar et al. \cite{11694284_Zargar} recommended a linear state estimation process by linearizing the AC power flow equations, assuming linear correlation between the measurements and the state variables. 

The DSSE solvers presented in the above-mentioned references mainly rely on physics-based models of the distribution networks, which frame a computationally expensive non-convex optimization problem. The following are the challenges for the application of optimization-based DSSE tools:
\begin{enumerate}
    \item The DSSE tools in \cite{336098_Baran}--\cite{7898803_Nanchian}, \cite{11370943_Xu}, are primarily developed by incorporating non-convex AC power-flow equations. These solvers are numerically robust but must be re-run in full for every new load scenario. Further, the methods adopt an iterative solution process, Gauss-Newton or Ordinal, because of the non-convexity. These techniques may suffer from the non-convergence problem in DSs \cite{Abur2004}. 
    \item To avoid the non-linear expressions, Taylor series expansion is adopted in \cite{9773346_Soltani} and \cite{9432710_Khan}. The linear set of equations is solved iteratively, but the convergence of these algorithms can only be achieved if the initialization is done close to the actual solution point. 
    \item Non-iterative linear DSSE solvers are proposed in \cite{11370943_Xu} and \cite{11694284_Zargar}. However, they assume a smooth change in the measurements and linear correlation between the measurements and the state variables, respectively, which may not hold under practical distribution network operating conditions. 
\end{enumerate}

\subsubsection{\textbf{Learning-Based DSSE Tools}}
By leveraging the wealth of real-time and historical data, data-driven approaches offer significant potential to enhance monitoring accuracy and improve the operational performance of DSs. In this regard, \cite{10963003_Pandey} and \cite{6176289_Manitsas} estimated bus injections from real-time measurements using an Artificial Neural Network (ANN). Previously, the authors also carried out a detailed comparative analysis in \cite{11453843_Kumar} between ANN and four other data-driven approaches (viz. Random Forest (RF), k-Nearest Neighbors (k-NN), Long Short-Term Memory (LSTM), and Convolutional Neural Networks (CNN)) and reported that ANN outperforms the others in estimating the state variables and modeling the non-linear relationship between the state variables and the measurements. 

To capture the intricate dependencies in distribution network measurements and improve state estimation performance, deep learning modules are employed in \cite{10374418_Wang, 11036748_Rout, 9910008_Wu, 10824822_Liu} to learn more complex features. Wang et al. \cite{10374418_Wang} and Rout and Natarajan \cite{11036748_Rout} proposed model-free attention-enhanced recurrent neural network (A-RNN) and data-driven deep unfolding learning approaches as DSSE tools. An Unrolled Spatiotemporal Graph Convolutional Network (USGCN) is suggested in \cite{9910008_Wu} to synchronously capture the spatiotemporal correlations between the long-range measured data. Liu et al. \cite{10824822_Liu} developed a model-free Generative Adversarial Network (GAN) to comprehend the correlation between the local measurements and the system states. 


Recent advances in learning-based DSSE are particularly focusing on merging data-driven learning modules with conventional optimization-based approaches and physical models of the DSs to improve estimation accuracy and robustness. In \cite{8693907_Zamzam}, the authors proposed a hybrid approach where a deep neural network (DNN) is utilized to provide initial solutions to the Gauss-Newton-based iterative DSSE algorithm. A hierarchical DSSE algorithm is introduced in \cite{9726921_Yuan}, where deep actor-critic reinforcement learning (DACRL), initialized by the WLS-estimated primary network voltages, is used to derive the states of secondary distribution network. Similarly, Zhang et al. \cite{11364252_Zhang} initialized the offline training of DNN by solving three different conventional DSSE algorithms, viz. WLS, weighted least absolute value (WLAV), and quadratic linear (QL). Zamzam and Sidiropoulos \cite{9072507_Zamzam_Nicholas} proposed a physics-aware neural network (PAWNN) by incorporating the WLS formulation into the graph neural network (GNN). Another DARCL-based algorithm is suggested in \cite{11603873_Chen}, where deep Kalman filter is implemented to capture the temporal correlation between the measurements at different time steps. Aiming to build a physics-aware graph learning, Wu et al. \cite{10836912_Wu} developed Meta-learning-enhanced physics-informed graph attention convolutional network (Meta-PIGACN), where Gauss-Newton based iterative solution process and bus admittance matrix elements are incorporated into the PIGACN layers through element-wise multiplication and admittance-based weight masking, respectively. Cao et al. \cite{10143253_Cao} combined the GAN with Bayesian Probability Weighted Averaging (BPWA) approach, where the GAN is initially trained to embed the topologies of a network with known states and then its weights are adjusted with BPWA to estimate the state variables of that network with different unseen topology. Reference \cite{10530443_Hu} merged GAN with Gaussian process to estimate the state variables in decentralized manner by adopting graphical representation of the measurements. To capture both global and local spatial dependencies on the measurements, Kim et al. \cite{11455366_Kim} proposed a structure-aware recurrent neural network (SARNN) integrated with an autoregressive moving average GNN (ARMA-GNN).

Despite the significant number of research studies, the existing learning-based DSSE tools exhibit following notable limitations: 
\begin{enumerate}
    \item The black box type learning-based DSSE tools, \cite{10963003_Pandey}--\cite{10824822_Liu}, learn the non-linear mapping relationship between the state variables and the recorded measurements by minimizing a pre-specified loss function over the training data. However, the absence of the DS physical model decreases the robustness and increases the chance of overfitting in the learning model \cite{9743327_Huang}. 
    \item The studies in \cite{8693907_Zamzam}--\cite{11364252_Zhang} either used learning-modules to provide initial solutions to the optimization-based tools or vice versa. However, for both the cases, the proposed algorithm involves solving a highly non-convex optimization problem.
    \item Reference \cite{9072507_Zamzam_Nicholas}--\cite{11455366_Kim} proposed grey box learning modules by embedding DS physical models into the learning models to enhance interpretability. However, their capability is generally limited to the trained network configurations, and fails to provide promising results to unseen networks. Although, Kim et al. \cite{11455366_Kim} proved transferability of their model to an unseen test network but the size of the training and testing networks are nearly same. In particular, the ability to perform zero-shot cross-grid transfer, i.e., directly applying a model trained on a small network to a larger, unseen target network without retraining or fine-tuning, remains insufficiently explored. Addressing this limitation would eliminate the need for separate model training for individual networks and improve the scalability and practical applicability of learning-based DSSE methods.   
\end{enumerate}

\subsection{Research Contributions}
It is evident from \cite{9072507_Zamzam_Nicholas}--\cite{11455366_Kim} that the graph based learning, particularly GNNs, graph convolutional networks (GCNs), and graph attention networks (GATs), provide a natural framework to incorporate network physics into the learning process. Graph approach helps to capture the spatial correlation from neighboring nodes \cite{Kipf2016} and integrate network AC power flow constraints to enable more physics-aware state estimation. Despite of the high computational efficiency, the existing approaches lack the following key attributes in terms of zero-shot cross-grid transfer.
\begin{enumerate}
    \item The primary limitation of GNN is structural rather than a matter of capacity or data volume. A GNN trained on a particular DS learns a function $f_{train}$ whose learned representations are validated only on the size and admittance regime seen in training. Applying $f_{train}$ to a differently sized graph does not merely reduce the estimation accuracy; for architectures with node-indexed parameters, such transfer is dimensionally undefined. Therefore, it suffers from topology lock-in, operational fragility (a single switching action alters the adjacency matrix and can silently invalidate the model), and computational dead-end (scaling laws force training directly on expensive, large-bus-count simulations rather than allowing knowledge transfer from cheaper, smaller feeders).
    \item Graph based approaches generally aggregate features from only neighborhood using certain weights. However, in case of power networks, the node voltage magnitude and angles are not only influenced by their neighbors but also by global network characteristics, including the network topology and the substation bus voltage, which are not always noticed at neighboring vertex features. This affects the robustness of the models particularly during testing on large sized unseen networks.
    \item GAT excels in learning from adjacent nodes through attention mechanism, but their learned attention patterns are largely influenced by the specific training network topologies. Therefore, during zero-shot cross-grid testing with the previously learned associated attention weights to unseen large networks, its performance degrades \cite{11455366_Kim}.
\end{enumerate}




In this regard, this article proposes an inductive physics-informed graph neural operator (IP-GNO), inspired from \cite{Li2020NeuralOperator}, that casts voltage estimation as operator learning, derived from the Neumann-series expansion of the inverse bus admittance matrix ($Y_{bus}^{-1}$). It combines an admittance-weighted kernel, a self-admittance gate on the zeroth-order term, a persistent source-injection skip connection, a five-feature positional encoding built from path impedance and a DC-linearized angle prior, and a sin/cos output head. The specific contributions are as follows:
\begin{enumerate}
    \item \textit{Admittance Operator:} From the Neumann series expansion of $Y_{\bus}^{-1}$, an admittance-parameterized kernel is derived to develop the graph neural operator applied to voltage estimation (both magnitude and angle). The admittance-parameterized kernel acts as a self-admittance gate on the zeroth-order term, and a persistent source-injection skip connection carried through every layer. An ablation study shows that extending the gate to every layer is redundant. 
    \item \textit{Physics-Embedded Tree Encoding (PETE):} Physics-Embedded Tree Encoding (PETE) is introduced as five positional node features derived from bus-injection-model path impedance and a DC power-flow angle prior, grid-size invariant by construction and distilled, via ablation, from an initial six-feature design.
    \item \textit{Sin/Cos Output Head:} The loss function of the training process is specifically designed to enforce accuracy across both physical dimensions of complex bus voltages while overcoming phase angle periodicity. By pairing a standard squared $L_2$ norm for voltage magnitudes with a trigonometric sine-cosine embedding for phase angles, it effectively measures the squared chordal distance on the unit circle. This dual formulation prevents artificial gradient spikes and numerical discontinuities at the $\pm\pi$ boundary, providing a smooth, continuous objective that maintains stable neural network training while precisely capturing grid voltage profiles and power-flow dynamics.
    \item \textit{Validation and Zero-Shot Cross-Grid Transfer:} The proposed learning module is validated on two small (IEEE 33 and IEEE 69-bus), two medium (IEEE 123 and IEEE 136-bus), and one large (2522-bus, reduced version of IEEE 8500-bus) sized test systems. Further, to check the efficacy for zero-shot cross-grid transfer capability, the IP-GNO module is trained on IEEE 33-bus system and then tested on other four distribution feeders, considered as unseen test network. The superiority of the proposed technique is established by performing a detailed comparative analysis with four other graph learning-based DSSE tools, viz. VanillaGNO, GAT, GINE (Graph Isomorphism Network with Edges), and GraphSAGE.
\end{enumerate}

\begin{figure*}[!t]
\centering
\includegraphics[width=\textwidth]{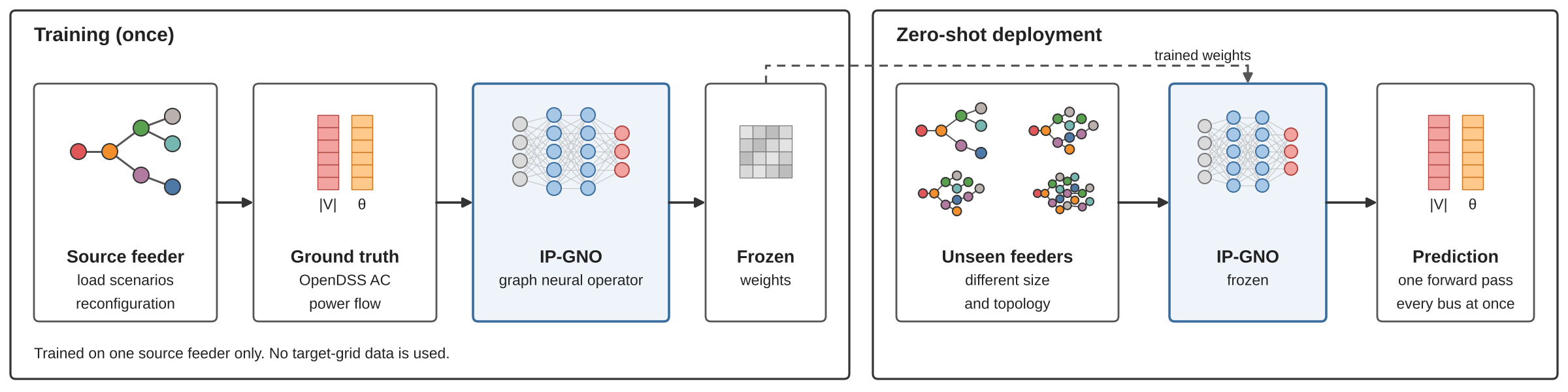}
\caption{Overview of IP-GNO. A single model is trained once on a small source feeder against OpenDSS ground truth, then frozen and applied zero-shot, in one forward pass, to unseen radial feeders of different size.}
\label{fig:overview}
\end{figure*}

\begin{figure*}[!t]
\centering
\includegraphics[width=0.96\textwidth]{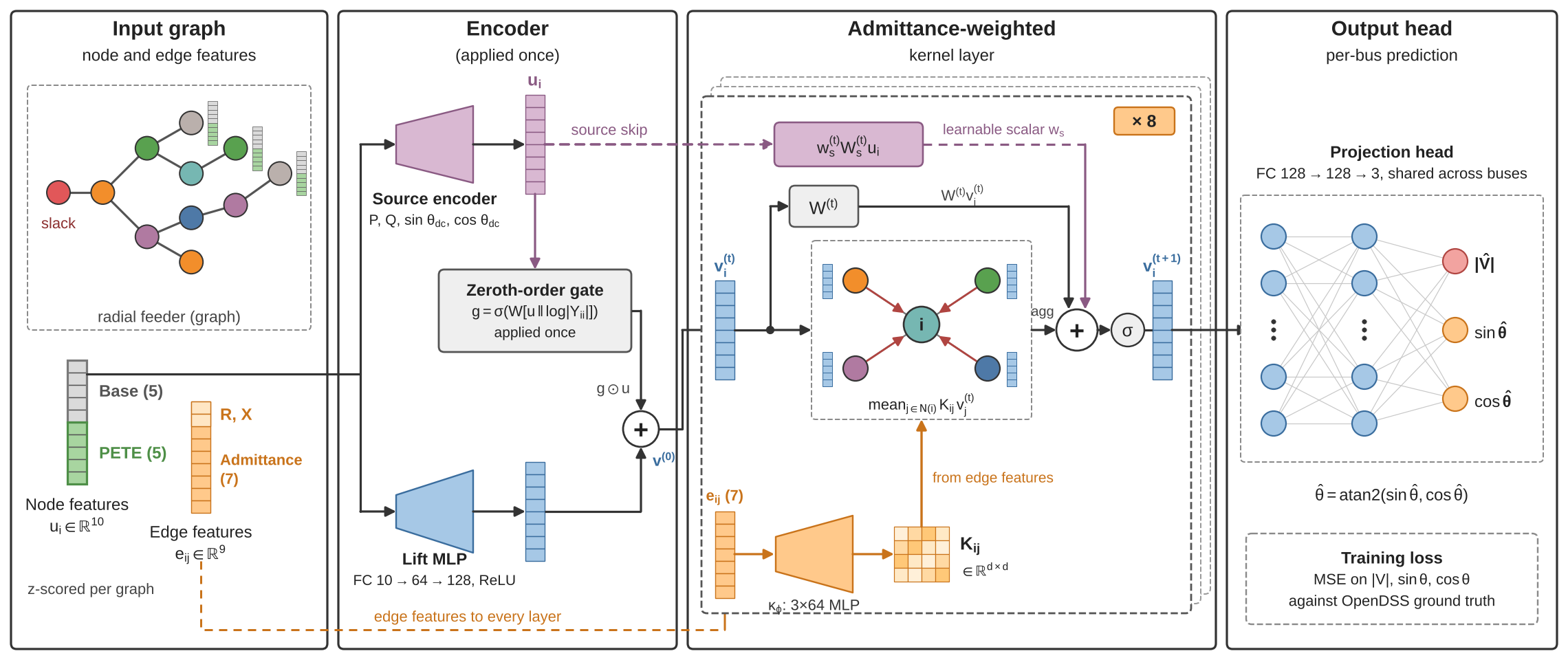}
\caption{IP-GNO architecture. The 10-dimensional node features (five base, five PETE) are lifted to $d=128$. The self-admittance gate is applied once, to the encoded source $u_i$, to form $v^{(0)}$. Each of the $T=8$ kernel layers applies a kernel conditioned on the seven admittance features of the 9-dimensional edge vector, aggregates by neighbourhood mean, and re-injects $u_i$ through a learnable-strength source skip. A projection head outputs $|\hat V|$, $\sin\hat\theta$, and $\cos\hat\theta$. A per-layer gate was tested and removed (Section~\ref{sec:kernel}).}
\label{fig:arch}
\end{figure*}

\section{Mathematical Foundations for the Proposed Admittance Operator}
\label{sec:neumann}

Consider a distribution network with $n$ buses, indexed by $i$ and $j$, where the series impedance of the line joining buses $i$ and $j$ is $z_{ij}=r_{ij}+jx_{ij}$ and its admittance is $y_{ij}=1/z_{ij}$. The diagonal and off-diagonal elements of the bus admittance matrix $\mathbf{Y}_{\bus}$ are
\begin{equation}
Y_{ii} = y_i^{sh} + \sum_{j \in \mathcal{N}(i)} y_{ij}, \qquad Y_{ij} = -y_{ij}\ \ (i\neq j),
\end{equation}
where $y_i^{sh}$ is the shunt admittance at bus $i$ and $\mathcal{N}(i)$ is the set of buses connected to bus $i$. The nodal voltages and current injections are related by Kirchhoff's current law,
\begin{equation}
\mathbf{Y}_{\bus}\mathbf{V}=\mathbf{I}_{\inj},
\label{eq:kcl}
\end{equation}
where $\mathbf{I}_{\inj}=[I_1,\ldots,I_n]^T$ and $\mathbf{V}=[V_1,\ldots,V_n]^T$ with $V_i=|V_i|\angle\theta_i$. The slack bus is eliminated from \eqref{eq:kcl}, so that its fixed voltage enters only through the injections of the buses adjacent to it. The injections are not specified directly. A load or generator specifies a complex power $S_i=P_i+jQ_i$, and the current $I_i=(S_i/V_i)^{*}$ depends on the unknown voltage, which is the nonlinearity that iterative power flow solvers resolve. In the linearized regime, where $\mathbf{I}_{\inj}$ is treated as the known source, \eqref{eq:kcl} gives $\mathbf{V}=\mathbf{Z}_{\bus}\mathbf{I}_{\inj}$ with $\mathbf{Z}_{\bus}=\mathbf{Y}_{\bus}^{-1}$.

This relation has the same form as the solution of a linear boundary value problem. For a PDE with differential operator $\mathcal{L}$, the solution is $u(x)=\int_\Omega G(x,y)f(y)\,dy$, where $G$ is the Green's function of $\mathcal{L}$. The matrix $\mathbf{Y}_{\bus}$ is a complex-weighted graph Laplacian, which is the discrete counterpart of $\mathcal{L}$, and $G(i,j)\equiv[\mathbf{Y}_{\bus}^{-1}]_{ij}$ gives the change in voltage at bus $i$ per unit current injected at bus $j$. The quantity to be learned is therefore a solution operator
\begin{equation}
\mathbf{V}=\mathcal{G}(\mathbf{S}_{\inj};\,\mathbf{Y}_{\bus}),
\end{equation}
which maps the injections to the voltages and is parameterized by the topology and the line admittances. This is the object targeted by the neural operator literature, namely the Graph Neural Operator (GNO) and the Fourier Neural Operator (FNO) of Li \emph{et al.} \cite{Li2020NeuralOperator, li2020fno}, in which a solution operator is learned independently of the mesh used to discretize the domain. The framing also explains why a standard GNN fails to transfer across grid sizes: it fits a function tied to the feeder it was trained on, for example $f_{33}$, and not the operator $\mathcal{G}(\cdot\,;\mathbf{Y}_{\bus})$.

To see how such an operator can be realized by message passing on the feeder graph, we expand $\mathbf{Y}_{\bus}^{-1}$ explicitly. Write $\mathbf{Y}_{\bus}=D+O$, where $D=\operatorname{diag}(Y_{11},\ldots,Y_{nn})$ collects the self-admittances and $O=\mathbf{Y}_{\bus}-D$ holds the off-diagonal mutual admittances, $O_{ij}=Y_{ij}=-y_{ij}$. If the spectral radius satisfies $\rho(D^{-1}O)<1$, which is the standard Jacobi convergence condition and is assumed here for the radial feeders considered, the Neumann series gives
\begin{equation}
\begin{split}
\mathbf{Y}_{\bus}^{-1}&=\sum_{k=0}^{\infty}\big(-D^{-1}O\big)^k D^{-1}\\
&= D^{-1}-D^{-1}OD^{-1}+D^{-1}OD^{-1}OD^{-1}-\cdots,
\end{split}
\label{eq:neumann}
\end{equation}
and therefore
\begin{equation}
\mathbf{V}=\underbrace{D^{-1}\mathbf{I}_{\inj}}_{\text{0-hop, self term}}\ \underbrace{-\,D^{-1}OD^{-1}\mathbf{I}_{\inj}}_{\text{1-hop correction}}\ +\ \cdots .
\end{equation}
The $k$-th term involves $k$ products of the off-diagonal matrix $O$, so it couples bus $i$ only to buses that are $k$ hops away on the graph. Truncating the series after $K$ terms is therefore equivalent to propagating information over a $K$-hop neighbourhood. The same truncation can be written as a recursion. Defining $\mathbf{V}^{(0)}=D^{-1}\mathbf{I}_{\inj}$ and
\begin{equation}
\begin{split}
\mathbf{V}^{(k+1)}&=D^{-1}\mathbf{I}_{\inj}-D^{-1}O\,\mathbf{V}^{(k)}\\
\Longleftrightarrow\;
V_i^{(k+1)}&=\frac{1}{Y_{ii}}\Big(I_i-\sum_{j\in\mathcal{N}(i)}Y_{ij}\,V_j^{(k)}\Big),
\end{split}
\label{eq:jacobi}
\end{equation}
the iterate $\mathbf{V}^{(k)}$ equals the sum of the first $k+1$ terms of \eqref{eq:neumann}. This is the Jacobi iteration, and it has a specific structure. Each update at bus $i$ uses only the current values at its neighbours, weighted by the admittance of the connecting line and scaled by the self-admittance at bus $i$, and the source $I_i$ is added again at every step. In other words, the truncated Neumann series is a message passing scheme on the network graph with a fixed, physically determined kernel.

This observation is what links the series to the GNO. Li \emph{et al.} \cite{Li2020NeuralOperator} approximate the action of a kernel integral operator by an iterative update in which each node aggregates the states of its neighbours through a kernel that depends on the edge between them,
\begin{equation}
\begin{split}
v^{(t+1)}(x) = \sigma\!\Bigg( & W^{(t)} v^{(t)}(x) \\
 & + \frac{1}{|\mathcal{N}(x)|}\sum_{y \in \mathcal{N}(x)} \kappa_\phi^{(t)}\big(e(x,y)\big)\, v^{(t)}(y) \Bigg),
\end{split}
\label{eq:gno}
\end{equation}
where $e(x,y)$ is the edge feature, $W^{(t)}$ is a linear map acting on the node's own state, $\sigma$ is a nonlinearity, and $\kappa_\phi^{(t)}$ maps the edge feature to a $d\times d$ mixing matrix. Comparing \eqref{eq:jacobi} with \eqref{eq:gno} shows that the two updates have the same form. In both, a node combines its own state with a sum over neighbours of an edge-dependent operator applied to the neighbour state. The Jacobi kernel $-Y_{ij}/Y_{ii}$ is a fixed scalar function of the line admittance, whereas $\kappa_\phi$ is a learned function of the edge features, so the neighbour term of \eqref{eq:jacobi} is a special case of the neighbour term of \eqref{eq:gno}. Each layer of the network corresponds to one Jacobi step, so a network with $T$ layers has the same $T$-hop receptive field as the series truncated at order $T$. The GNO then generalizes the series in three ways that suit the present problem. The scalar coefficient becomes a $d\times d$ matrix acting on a lifted state of dimension $d>1$, the kernel is learned from data and so absorbs the degree normalization $1/|\mathcal{N}(x)|$ and the scaling by $Y_{ii}$ that the exact coefficient would require, and the nonlinearity $\sigma$ allows the network to represent the dependence of $I_i$ on $V_i$ and the truncation error, neither of which the linear series captures. The correspondence is therefore structural and not an exact identity, and it is the reason a GNO is a natural and not an arbitrary choice for estimating voltages in a network.

The structure of \eqref{eq:neumann} and \eqref{eq:jacobi} also fixes the design requirements for the operator. First, every term is built from products of $Y_{ij}$ and $D^{-1}$, so a kernel that receives admittance-derived quantities can represent each term directly, while a kernel that receives only $(R_{ij},X_{ij})$ must additionally learn the nonlinear map $(R,X)\mapsto Y=1/(R+jX)$, which increases the approximation burden. Second, the factor $D^{-1}$ multiplies every term. Section~\ref{sec:kernel} shows that this is most effectively captured as an explicit gate on the zeroth-order term alone, since a gate inside every later layer is redundant, most likely because the residual and skip connections already carry that information forward. Third, $\mathbf{I}_{\inj}$ appears in every term and is re-added at every step of \eqref{eq:jacobi}. If it were encoded only at the input layer, deeper layers would lose access to it through the saturating nonlinearities, so it must be re-injected at each layer. Finally, the kernel $\kappa_\phi$ acts on edge features and its weights are shared across all nodes, so its domain does not depend on node indices or on the number of buses. This is the topology and resolution invariance of the GNO that is exploited in Section~\ref{sec:zeroshot}.

\section{Proposed Methodology: IP-GNO}
\label{sec:method}

\subsection{Overview}

IP-GNO instantiates the GNO framework with a kernel and auxiliary structures aligned with the Neumann-series decomposition of Section~\ref{sec:neumann}. A 10-dimensional node feature vector and a 9-dimensional edge feature vector are lifted to a $d=128$-dimensional latent space. A self-admittance gate is applied once, to the zeroth-order term, before any message passing; the result is then passed through $T=8$ admittance-weighted kernel layers, each carrying a source-injection skip connection (not self-admittance-gated), and projected to a 3-channel output $(|\hat V|, \sin\hat\theta, \cos\hat\theta)$. Training uses \textbf{full graphs} of the training feeder, not extracted subgraphs.

Fig.~\ref{fig:overview} summarizes the train-once, deploy-zero-shot workflow and Fig.~\ref{fig:arch} the architecture. ``Physics-informed'' refers to the architecture and the input features, which are derived from the Neumann expansion and from path-impedance and DC-linearized quantities; no physics-residual term is added to the training loss (Eq.~\eqref{eq:loss}).

\subsection{Node and Edge Feature Design}

\textbf{Node features} ($u \in \mathbb{R}^{10}$): \texttt{is\_slack}, \texttt{has\_load}, a per-graph $z$-scored $P_{pu}$ and $Q_{pu}$, and a per-graph $z$-scored log self-admittance $\widetilde{\log|Y_{ii}|}$ (5 base features), concatenated with the five PETE positional features of Section~\ref{sec:pete}.

\textbf{Edge features} ($e \in \mathbb{R}^{9}$): the raw series impedance $(R_{pu}, X_{pu})$, together with seven admittance-derived quantities used by the kernel: $|Y|,\ \angle Y,\ G,\ B,\ \log|Y|,\ \log G,\ \log|B|$, where $Y_{ij}=1/z_{ij}=G_{ij}+jB_{ij}$.

\textbf{Per-graph edge normalization.} All nine edge columns are $z$-score normalized \emph{per graph}, not globally, immediately before the kernel sees them. Because the kernel only ever sees relative admittance patterns within a graph, the statistical distribution presented to it is invariant to grid size and conductor type, even though the absolute admittances of a 33-bus and a 2522-bus feeder differ by orders of magnitude.

\subsection{PETE: Physics-Embedded Tree Encoding}
\label{sec:pete}

Standard GNN positional encodings -- Laplacian eigenvectors, random-walk features \cite{dwivedi2022} -- capture only graph connectivity and symmetry, which is insufficient on a radial feeder: two degree-1 leaf buses can be structurally indistinguishable to such an encoding while sitting at very different electrical distances from the substation. PETE was initially designed around six positional features, computed via a single BFS traversal of the radial tree rooted at the slack bus:
\begin{align}
\text{x\_path\_pu}[i] &= \frac{\sum_{k \in \text{path}(0\rightarrow i)} X_k}{\text{scale}_X}, \\
\text{b\_path\_pu}[i] &= \frac{\sum_{k \in \text{path}(0\rightarrow i)} B_k}{\text{scale}_B}, \\
\text{downstream\_load\_frac}[i] &= \frac{\text{load downstream of } i}{\text{total feeder load}}, \\
\text{degree\_clamp3}[i] &= \min\!\Big(\tfrac{\deg(i)}{3},\,1\Big), \\
\big(\sin\theta_{dc}[i],\ \cos\theta_{dc}[i]\big) &: \quad B'\theta_{dc} \approx P,
\end{align}
where the last pair is the sine and cosine of the classical DC power-flow angle solution \cite{stott2009} -- a bus-injection-model linearization solving $B'\theta \approx P$ for the susceptance matrix $B'$ assembled from line reactances, giving each bus a closed-form angle prior before any message passing occurs. \texttt{x\_path\_pu} and \texttt{b\_path\_pu} are the (per-graph-normalized) reactive and susceptance-adjacent parts of the diagonal path impedance to the slack, following the tree path-sum identity $Z_{ii} = \sum_{k\in\text{path}(0\to i)} z_k$; \texttt{downstream\_load\_frac} summarizes the aggregated downstream load a bus's own line segment must carry.

\begin{figure*}[!t]
\centering
\includegraphics[width=0.92\textwidth]{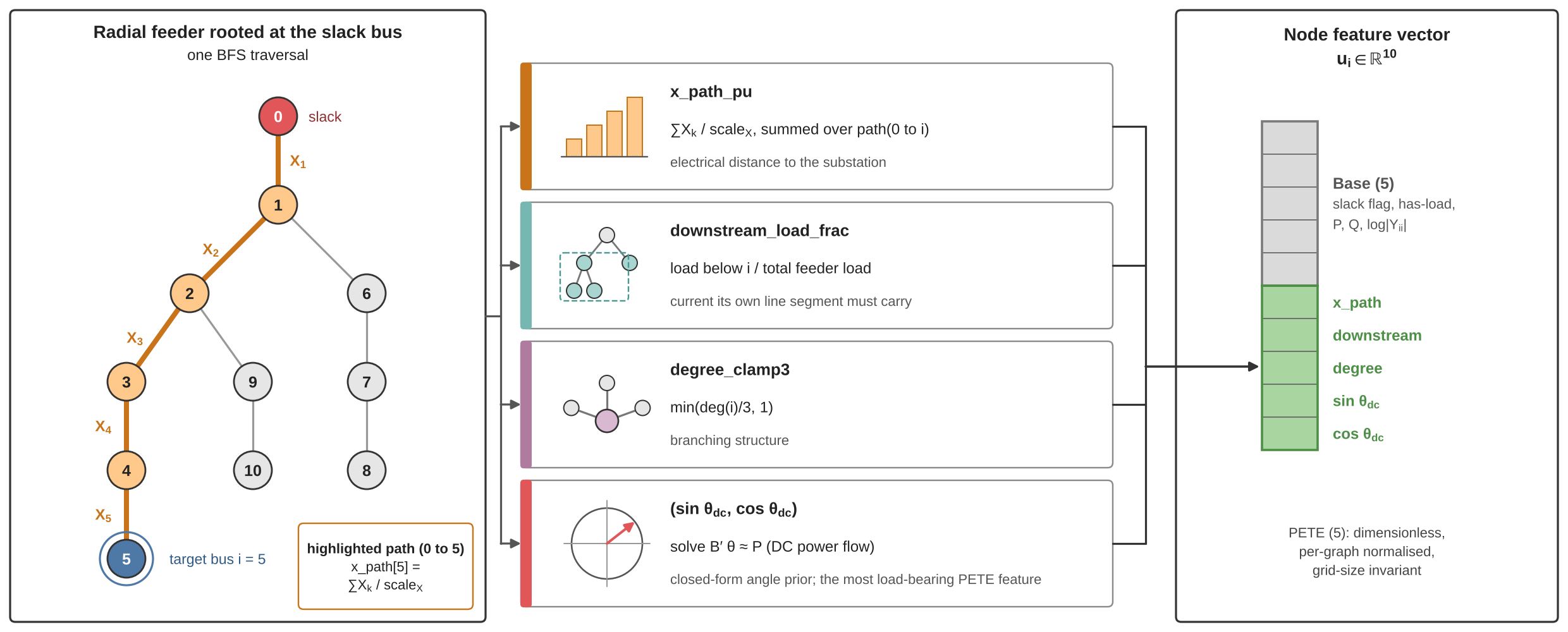}
\caption{PETE on a toy radial feeder rooted at slack bus~0. The path-sum feature accumulates series reactance along the unique path from the slack bus (highlighted for bus~5); the downstream-load fraction, clamped degree, and DC-angle prior are computed from the same rooted tree. Together with the five base features they form the 10-dimensional node vector. All features are dimensionless and normalized per graph, so they do not depend on feeder size.}
\label{fig:pete}
\end{figure*}

Fig.~\ref{fig:pete} shows how the features are obtained from a single BFS traversal of a small radial feeder.

\textbf{The final architecture uses five of these six features, dropping \texttt{b\_path\_pu}.} The ablation (Section~\ref{sec:ablation}) shows that the PETE features other than the DC angle prior contribute only marginally when removed individually, and \texttt{b\_path\_pu} appears redundant with \texttt{x\_path\_pu}, which carries the dominant reactive-path signal and is retained. The DC angle prior, by contrast, is the single most load-bearing PETE feature by a wide margin.

PETE and per-graph normalization are computed once per graph, on that graph's true topology and true slack bus.

\subsection{Admittance-Weighted Kernel Layer and Self-Admittance Gate}
\label{sec:kernel}

\begin{proposition}
\label{prop:kernel}
A GNO kernel taking $(|Y|, \angle Y, G, B, \log|Y|, \log G, \log|B|)$ as input, combined with a self-admittance gate on the zeroth-order term and a source-injection skip connection carried through every layer, can represent the terms of the Neumann expansion of $Y_{\bus}^{-1}$ directly from its inputs, whereas an otherwise identical kernel using only $(R,X)$ and no gate or skip must additionally approximate the map $(R,X)\mapsto(G,B)$ and has no persistent access to the source term.
\end{proposition}

\emph{Proof sketch.} Let $\kappa^{*}$ denote the kernel that exactly reproduces the first-order Neumann term, $\kappa^{*}(e_{ij}) = -Y_{ij}D_j^{-1}$, a function of $Y_{ij}$ and $Y_{jj}$ alone. An MLP kernel $\kappa_\phi$ receiving $(G_{ij}, B_{ij}, Y_{jj})$ directly can approximate $\kappa^{*}$ to arbitrary precision as a function of those inputs; a kernel receiving only $(R_{ij}, X_{ij})$ must instead approximate the composition of the nonlinear inversion $(R,X)\mapsto(G,B)$ with $\kappa^{*}$, which increases the approximation burden at equal network capacity. The zeroth-order self-admittance gate mirrors the $D^{-1}$ prefactor of the leading Neumann term; the source skip keeps $I_{\inj}$ available at every layer.

Concretely, letting $u_i = \text{source\_enc}(P_i, Q_i, \sin\theta_{dc,i}, \cos\theta_{dc,i})$ denote the encoded injection (including the DC angle prior):
\begin{align}
v_i^{(0)} &= \text{lift}(x_i) + \underbrace{\sigma\big(W_{g}\,[u_i \,\|\, \widetilde{\log|Y_{ii}|}]\big)}_{\text{zeroth-order gate}} \odot\, u_i, \\
K_{ij} &= \kappa_\phi\big(e_{n,ij}[2{:}9]\big) \in \mathbb{R}^{d\times d}, \\
\text{agg}_i^{(t)} &= \frac{1}{|\mathcal{N}(i)|}\sum_{j\in\mathcal{N}(i)} K_{ij}\,v_j^{(t)}, \\
v_i^{(t+1)} &= \sigma\!\Big(W^{(t)} v_i^{(t)} + \text{agg}_i^{(t)} + w_s^{(t)}\, W_s^{(t)} u_i\Big),
\end{align}
for $t=0,\ldots,T-1$, where $\kappa_\phi$ is a 3-layer, 64-hidden-unit MLP and $w_s^{(t)}=\sigma(\text{logit}_s^{(t)})$ is a learnable per-layer scalar gate on the skip strength, initialized near zero. A \textbf{VanillaGNO} baseline uses an identical layer structure but replaces the admittance input to $\kappa_\phi$ with raw $(R_{pu}, X_{pu})$ and omits both the zeroth-order gate and the source skip.

\textbf{Gate placement.} Proposition~\ref{prop:kernel} motivates a self-admittance gate but does not require it at every layer. We compared a variant with an additional explicit gate at every kernel layer against the zeroth-order gate alone. On the 69-bus within-grid setting, removing the per-layer gate lowers $|V|$ MAE by 27.0\% and $\theta$ MAE by 51.9\% ($0.0822^{\circ}\to0.0395^{\circ}$). Since $\widetilde{\log|Y_{ii}|}$ is already a base node feature entering $v_i^{(0)}$ through the zeroth-order gate, and the residual pathways give every later layer access to it, the explicit per-layer gate is redundant rather than in conflict with the $D^{-1}$-prefactor argument. \textbf{IP-GNO's final architecture therefore applies the gate only at the zeroth-order term.}

\subsection{Sin/Cos Output Head}

Raw angle regression via $\text{atan2}$ implicitly treats an angle error near the $\pm180^{\circ}$ boundary as though the two values were $358^{\circ}$ apart, injecting a large false gradient for a near-zero true angular error. IP-GNO instead predicts $(|\hat V|, \sin\hat\theta, \cos\hat\theta)$ and trains with
\begin{equation}
\mathcal{L} = \big\lVert |V| - |\hat V| \big\rVert^2 + \lVert \sin\theta - \sin\hat\theta \rVert^2 + \lVert \cos\theta - \cos\hat\theta \rVert^2,
\label{eq:loss}
\end{equation}
recovering the angle at inference via $\hat\theta = \text{atan2}(\sin\hat\theta, \cos\hat\theta)$. Because $(\sin\theta,\cos\theta)\in\mathbb{R}^2$ has no periodicity discontinuity, this removes the wraparound pathology at no additional parameter cost. In a controlled ablation isolating this choice, sin/cos parameterization and direct angle regression perform comparably ($\theta$ MAE $0.0748^{\circ}$ vs.\ $0.0711^{\circ}$ within-grid, a difference within seed-to-seed variation); on the present grids and load ranges its measured effect is therefore modest, and it is retained as a safeguard against wraparound.

\subsection{Zero-Shot Topological Generalization}
\label{sec:zeroshot}

The kernel MLP's domain is edge-feature space $\mathbb{R}^7$, never node-index or graph-size space, and the same weights are applied at every node of every graph. IP-GNO is therefore defined on any radial feeder, which is the resolution-invariance property established for GNOs on PDE domains \cite{Li2020NeuralOperator}. Whether accuracy transfers under this invariance is an empirical question, addressed in Section~\ref{sec:crossgrid}: with no fine-tuning and no target-grid data, IP-GNO is the most accurate of five architectures on all four target grids, and only IP-GNO and VanillaGNO remain bounded and physically plausible on all four, while GAT and GINE diverge on several grids and GraphSAGE diverges on the 2522-bus network. Section~\ref{sec:withingrid} shows that IP-GNO also leads within-grid on all five feeders.

\section{Experimental Setup}
\label{sec:setup}

\begin{figure*}[!t]
\centering
\includegraphics[width=0.92\textwidth]{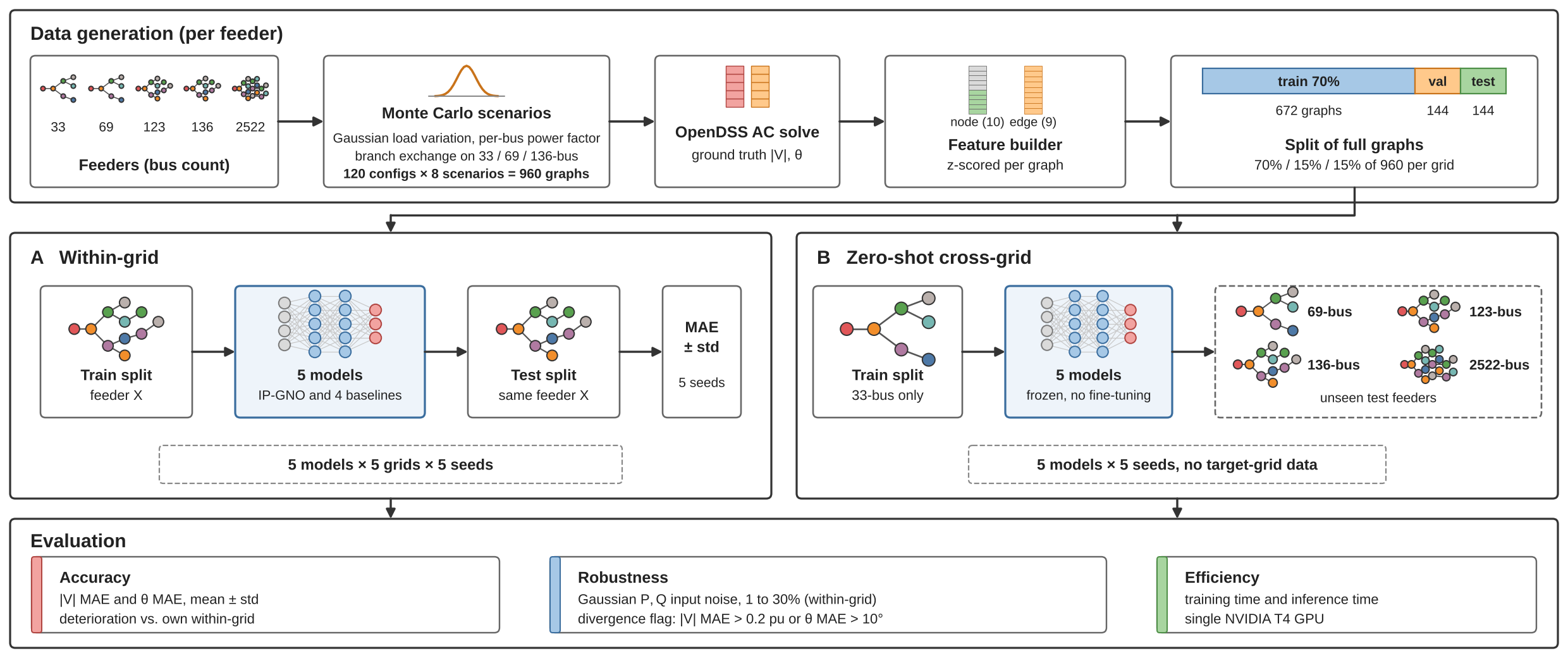}
\caption{Experimental protocol. Scenarios are generated per feeder and solved with OpenDSS; features are built per graph. Track~A trains and tests on the same feeder; Track~B trains on the 33-bus feeder only and tests on the four unseen feeders without fine-tuning.}
\label{fig:protocol}
\end{figure*}

Fig.~\ref{fig:protocol} summarizes the data-generation and evaluation protocol described below.

\textbf{Feeders.} IEEE 33-bus (training) and four zero-shot test feeders: IEEE 69-bus (Baran \& Wu \cite{baranwu1989}); the IEEE 123-bus feeder \cite{kersting1991}, positive-sequence-reduced from its native multi-phase representation, with its six voltage regulators modeled as fixed-ratio placeholder branches; the real Mantovani \emph{et al.} 136-bus system \cite{mantovani2000} (\texttt{case136ma}), which, unlike 123-bus, retains genuine reconfiguration switches; and a 2522-bus network that is a subsystem of the IEEE 8500-Node Test Feeder \cite{arritt2010}. AC ground truth is computed with the Newton--Raphson solution mode of OpenDSS \cite{dugan2011}. System base 1~MVA; each grid's own nominal voltage sets its per-unit base (12.66~kV for 33/69-bus; 4.16~kV for 123-bus; 13.8~kV for 136-bus; 7.2~kV for the 2522-bus network).

\textbf{Scenario generation and dataset size.} Load scenarios are generated by Monte Carlo sampling: Gaussian variation around each bus's base $(P,Q)$, with the per-bus power factor preserved and a fraction of scenarios drawn at elevated load to cover heavier operating points. For each grid, 120 configurations with 8 scenarios per configuration yield 960 full graphs, which are split exactly 70\% training (672 graphs), 15\% validation (144), and 15\% testing (144). For reconfigurable feeders (33, 69, 136), each configuration also draws an independent branch-exchange topology (close one tie switch, open one sectionalizing branch on the resulting loop); the 123-bus and 2522-bus networks have no reconfiguration data and use one fixed topology throughout, so load variation is their only source of scenario diversity. All feeders are treated as balanced, positive-sequence radial networks.

\textbf{Training protocol.} Full-graph training, validation, and testing on every feeder. Training is terminated by early stopping on validation $|V|$ MAE, so some runs end before the epoch cap.

\textbf{Hyperparameters.} $d=128$; $T=8$ kernel layers; kernel MLP 3 layers $\times$ 64 hidden units; Adam \cite{kingma2015}, learning rate $10^{-3}$, weight decay $10^{-4}$; batch size 32 (batch size 2 for the 2522-bus network); up to 150 epochs; dropout $p=0.1$. \textbf{Every result in Tables~\ref{tab:withingrid}--\ref{tab:deterioration} uses five training seeds per architecture per grid}, reported as mean $\pm$ standard deviation across seeds, with median-absolute-deviation-based outlier flagging (threshold $2.5\times1.4826\,\text{MAD}$) and clean-subset means reported.

\textbf{Baselines.} All baselines share the training protocol, hyperparameters, and PETE node features, isolating the specific architectural ablation each is designed to test (Table~\ref{tab:baselines}): GAT \cite{velickovic2018}, GINE \cite{hu2020,xu2019}, and GraphSAGE \cite{hamilton2017}, plus VanillaGNO. GINE (Graph Isomorphism Network with Edges) extends the GIN architecture \cite{xu2019} to incorporate edge features, such as line impedances, directly into the message-passing aggregation \cite{hu2020}.

\begin{table*}[t]
\centering
\caption{Compared architectures}
\label{tab:baselines}
\begin{tabular}{lccc}
\toprule
Model & Kernel input & Pos.\ features & Gate/skip \\
\midrule
VanillaGNO & $(R_{pu}, X_{pu})$ only & PETE & None \\
\textbf{IP-GNO} & 7 adm.\ features & PETE & Zeroth-order gate + skip \\
GAT & Multi-head attn. & PETE & -- \\
GINE & Edge-cond.\ GIN & PETE & -- \\
GraphSAGE & Neighborhood samp. & PETE, no edges & -- \\
\bottomrule
\end{tabular}
\end{table*}

\textbf{Metrics.} Voltage-magnitude and angle MAE (pu, deg); validation $|V|$ MAE versus epoch; training and inference time; percentage deterioration of $|V|$ MAE under Gaussian input noise.

\textbf{Computational setup.} All models were trained and evaluated on a single NVIDIA Tesla T4 GPU (16~GB) provided by Google Colab, using PyTorch and PyTorch Geometric. Ground-truth power flows were computed on the CPU with OpenDSS. Training IP-GNO (150 epochs) on the IEEE 33-bus system requires approximately 25 minutes, with a test-time inference speed of $\sim$8 milliseconds per batch.

\section{Results}

\begin{figure*}[!t]
\centering
\begin{minipage}[t]{0.325\textwidth}\centering
\figorplaceholder{convergence curves, 33-bus}{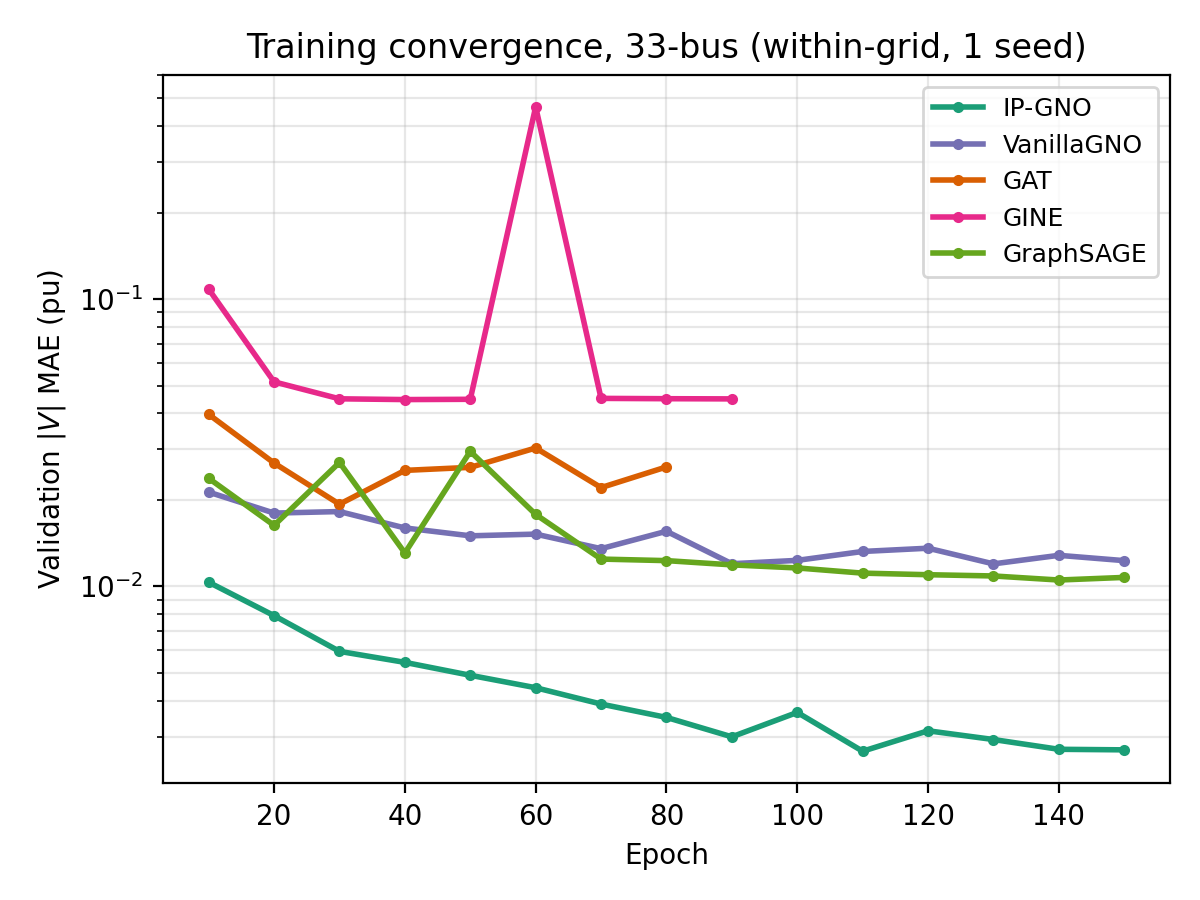}\\[-1pt]\footnotesize (a) 33-bus
\end{minipage}\hfill
\begin{minipage}[t]{0.325\textwidth}\centering
\figorplaceholder{convergence curves, 69-bus}{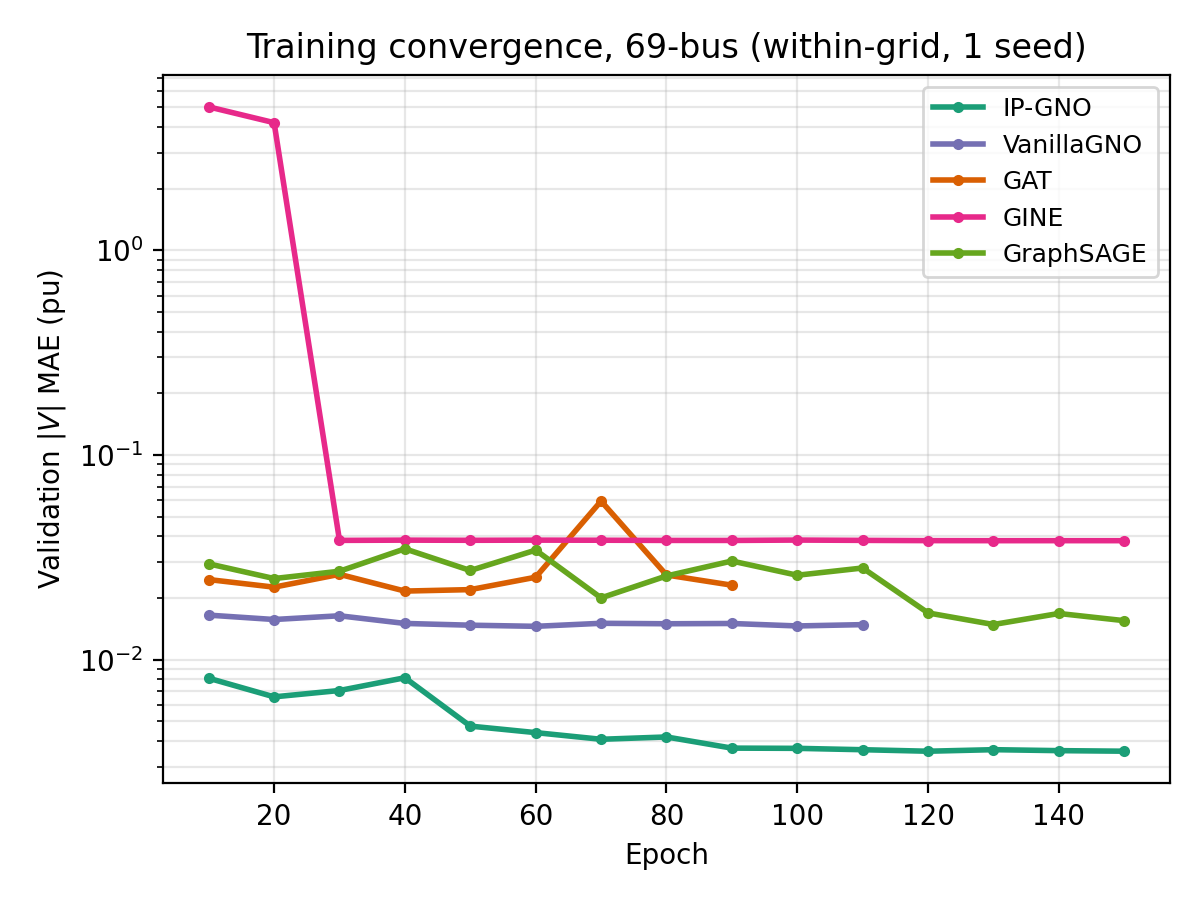}\\[-1pt]\footnotesize (b) 69-bus
\end{minipage}\hfill
\begin{minipage}[t]{0.325\textwidth}\centering
\figorplaceholder{convergence curves, 123-bus}{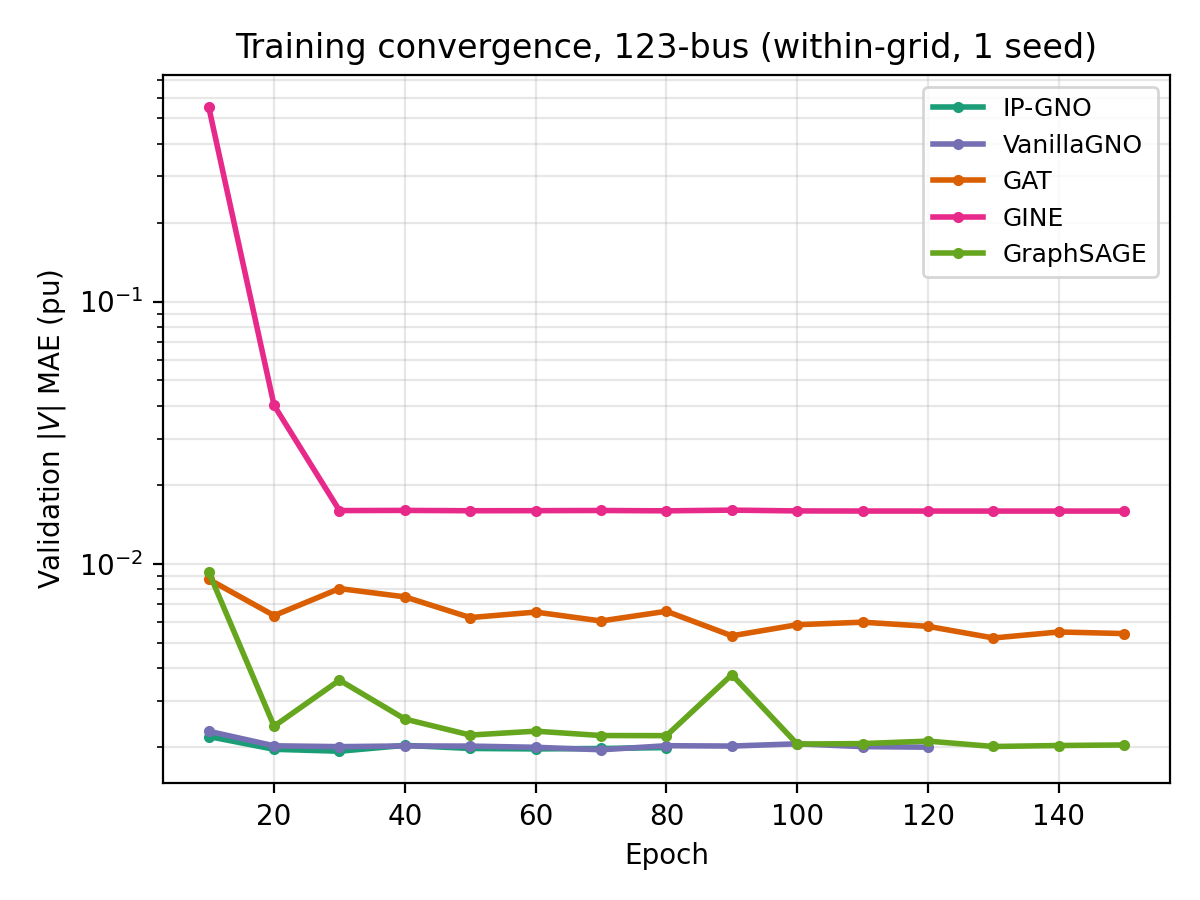}\\[-1pt]\footnotesize (c) 123-bus
\end{minipage}

\vspace{4pt}
\begin{minipage}[t]{0.325\textwidth}\centering
\figorplaceholder{convergence curves, 136-bus}{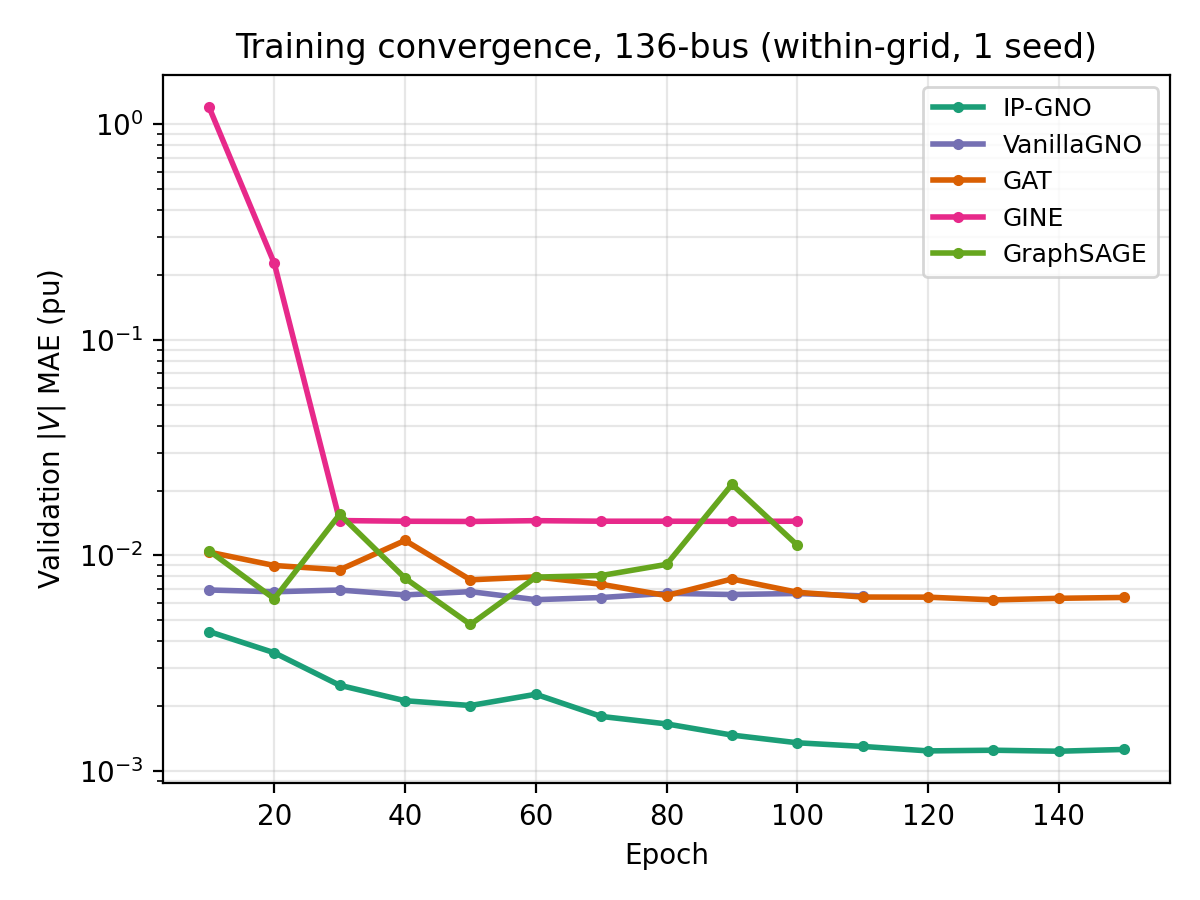}\\[-1pt]\footnotesize (d) 136-bus
\end{minipage}\hspace{0.02\textwidth}
\begin{minipage}[t]{0.325\textwidth}\centering
\figorplaceholder{convergence curves, 2522-bus}{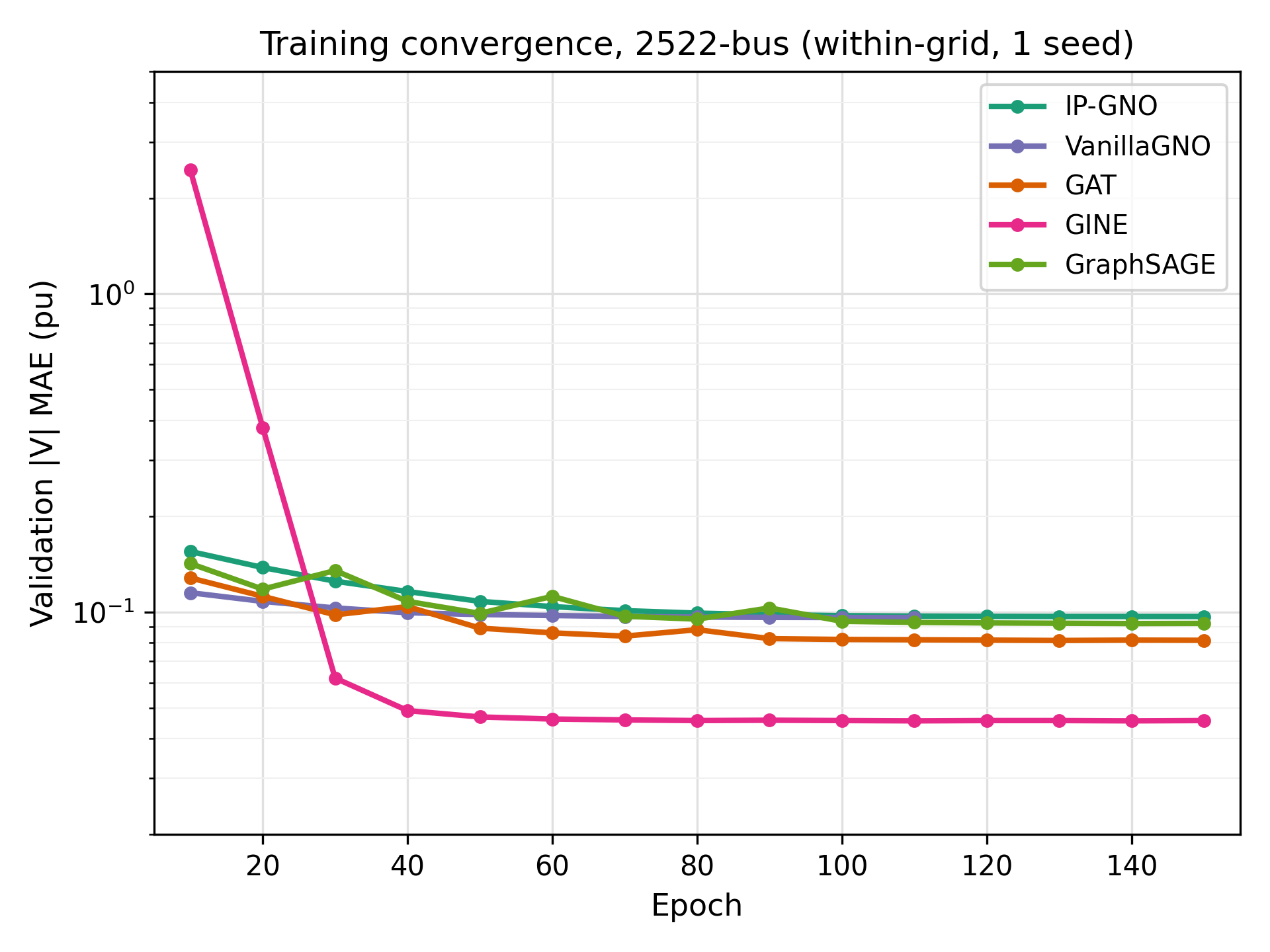}\\[-1pt]\footnotesize (e) 2522-bus
\end{minipage}
\caption{Training convergence of all five architectures: validation $|V|$ MAE (pu, log scale) versus epoch, sampled every 10 epochs, for within-grid training on each feeder (one seed). Curves that end before epoch 150 were terminated by early stopping. In (c), the IP-GNO and VanillaGNO curves nearly coincide, as does GraphSAGE from epoch 100.}
\label{fig:convergence}
\end{figure*}

\subsection{Within-Grid Evaluation}
\label{sec:withingrid}

Table~\ref{tab:withingrid} reports each architecture trained and tested on its own held-out split of the same grid, as mean $\pm$ standard deviation over five seeds, and Fig.~\ref{fig:convergence} shows the corresponding training convergence.

\begin{table*}[t]
\centering
\caption{Within-grid $|V|$ MAE (pu) and $\theta$ MAE (deg), train and test on the same feeder, mean $\pm$ standard deviation over 5 seeds. Bold = best per grid.}
\label{tab:withingrid}
\resizebox{\textwidth}{!}{
\begin{tabular}{l cc cc cc cc cc}
\toprule
\multirow{2}{*}{Grid} & \multicolumn{2}{c}{IP-GNO (Ours)} & \multicolumn{2}{c}{VanillaGNO} & \multicolumn{2}{c}{GAT} & \multicolumn{2}{c}{GINE} & \multicolumn{2}{c}{GraphSAGE} \\
\cmidrule(lr){2-3} \cmidrule(lr){4-5} \cmidrule(lr){6-7} \cmidrule(lr){8-9} \cmidrule(lr){10-11}
& $|V|$ ($\times 10^{-3}$) & $\theta$ (deg) & $|V|$ ($\times 10^{-3}$) & $\theta$ (deg) & $|V|$ ($\times 10^{-3}$) & $\theta$ (deg) & $|V|$ ($\times 10^{-3}$) & $\theta$ (deg) & $|V|$ ($\times 10^{-3}$) & $\theta$ (deg) \\
\midrule
33-bus & $\mathbf{2.30 \pm 0.12}$ & $\mathbf{0.045 \pm 0.002}$ & $7.95 \pm 0.81$ & $0.108 \pm 0.012$ & $12.7 \pm 0.95$ & $0.240 \pm 0.085$ & $32.5 \pm 1.15$ & $0.526 \pm 0.125$ & $7.66 \pm 0.52$ & $0.107 \pm 0.015$ \\
69-bus & $\mathbf{2.67 \pm 0.18}$ & $\mathbf{0.039 \pm 0.003}$ & $12.8 \pm 0.92$ & $0.150 \pm 0.021$ & $21.3 \pm 1.12$ & $0.280 \pm 0.092$ & $41.6 \pm 1.65$ & $0.610 \pm 0.142$ & $12.5 \pm 0.88$ & $0.190 \pm 0.025$ \\
123-bus & $\mathbf{1.88 \pm 0.09}$ & $\mathbf{0.058 \pm 0.004}$ & $2.88 \pm 0.25$ & $0.081 \pm 0.008$ & $4.57 \pm 0.65$ & $0.348 \pm 0.055$ & $11.7 \pm 0.35$ & $0.536 \pm 0.085$ & $1.92 \pm 0.15$ & $0.073 \pm 0.006$ \\
136-bus & $\mathbf{0.93 \pm 0.04}$ & $\mathbf{0.057 \pm 0.003}$ & $6.09 \pm 0.72$ & $0.293 \pm 0.045$ & $8.60 \pm 0.95$ & $0.431 \pm 0.065$ & $16.6 \pm 0.28$ & $0.821 \pm 0.115$ & $4.79 \pm 0.55$ & $0.232 \pm 0.035$ \\
2522-bus & $\mathbf{0.84 \pm 0.01}$ & $\mathbf{0.085 \pm 0.005}$ & $96.1 \pm 5.50$ & $5.59 \pm 0.85$ & $81.5 \pm 12.5$ & $5.36 \pm 1.05$ & $31.2 \pm 8.50$ & $3.12 \pm 0.75$ & $92.1 \pm 6.50$ & $5.35 \pm 0.82$ \\
\bottomrule
\end{tabular}
}
\end{table*}

IP-GNO leads on all five grids in both $|V|$ and $\theta$ MAE. In $|V|$ MAE it improves on the next-best baseline by 70.0\% (GraphSAGE) on 33-bus, 78.7\% (GraphSAGE) on 69-bus, 2.1\% (GraphSAGE) on 123-bus, 80.6\% (GraphSAGE) on 136-bus, and 97.3\% (GINE) on 2522-bus. The 123-bus margin is the narrowest and lies within seed-to-seed variation ($1.88\pm0.09$ vs.\ $1.92\pm0.15$), so the two models should be read as comparable there; 123-bus is also the only one of the reconfigurable-scale grids with no reconfiguration data (Section~\ref{sec:setup}), so every model is fit to a single fixed topology, which may leave less room for architectural differences to matter. On the largest grid, 2522-bus, IP-GNO's $|V|$ MAE is more than 37 times lower than that of the best baseline, while the four baselines remain between $3.12\times10^{-2}$ and $9.61\times10^{-2}$~pu.

\textbf{Training convergence.} On the 33-, 69-, 123-, and 136-bus feeders, IP-GNO's validation error falls steadily and flattens by roughly epoch 100--120, with only minor non-monotonic bumps; it lies below every baseline throughout on 33-, 69-, and 136-bus and is at or near the lowest on 123-bus, where it overlaps VanillaGNO and, later, GraphSAGE. GINE starts orders of magnitude higher (up to $\approx$5~pu on 69-bus) and settles on a flat plateau above IP-GNO's level, and GAT and GraphSAGE show intermittent excursions (e.g., GAT at epoch 70 on 69-bus, GraphSAGE at epochs 30 and 50 on 33-bus). Several curves end before epoch 150 because of early stopping (e.g., GINE and GraphSAGE at epoch 100 on 136-bus, GAT at epoch 90 on 69-bus). 
These curves are single-seed and complement, rather than replace, the five-seed results of Table~\ref{tab:withingrid}.

\subsection{Per-Bus Within-Grid Profiles}
\label{sec:withinprofiles}

\begin{figure*}[!tbp]
\centering
\begin{minipage}[t]{0.49\textwidth}\centering
\includegraphics[width=\linewidth]{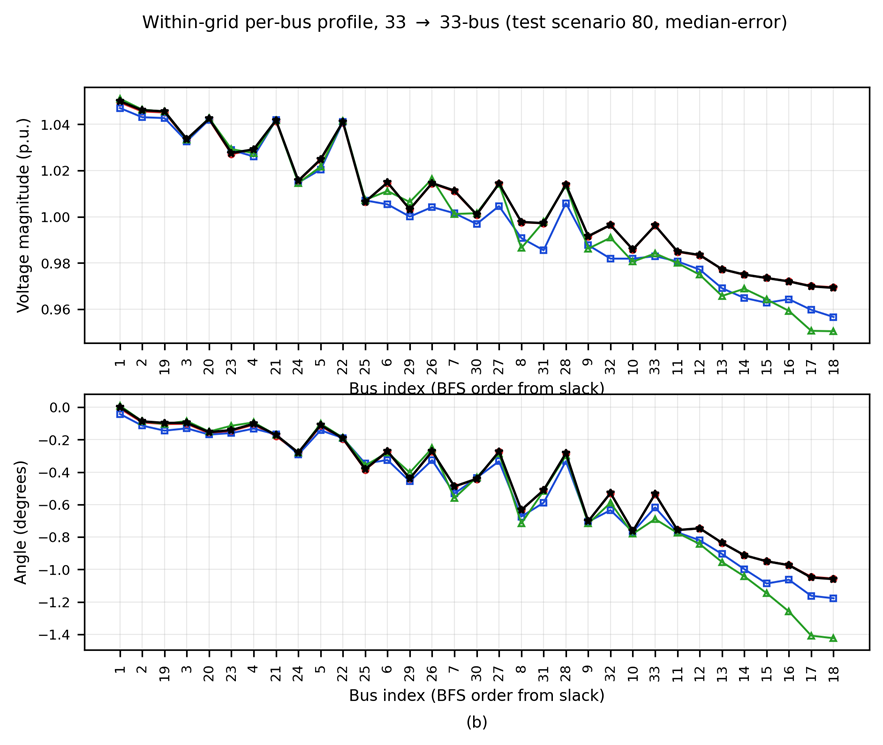}\\[-1pt]\footnotesize (a) 33-bus
\end{minipage}\hfill
\begin{minipage}[t]{0.49\textwidth}\centering
\includegraphics[width=\linewidth]{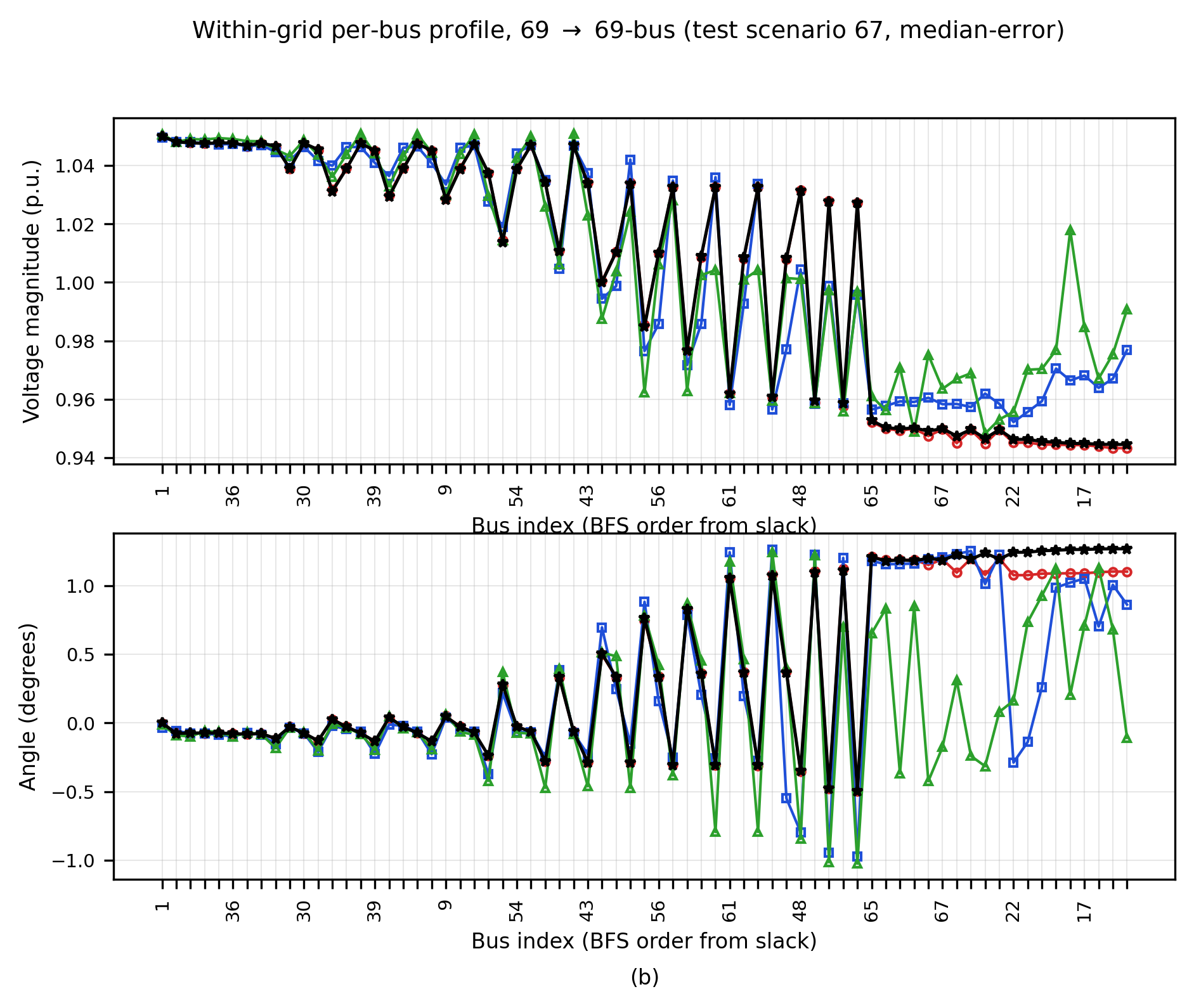}\\[-1pt]\footnotesize (b) 69-bus
\end{minipage}

\vspace{4pt}
\begin{minipage}[t]{0.49\textwidth}\centering
\includegraphics[width=\linewidth]{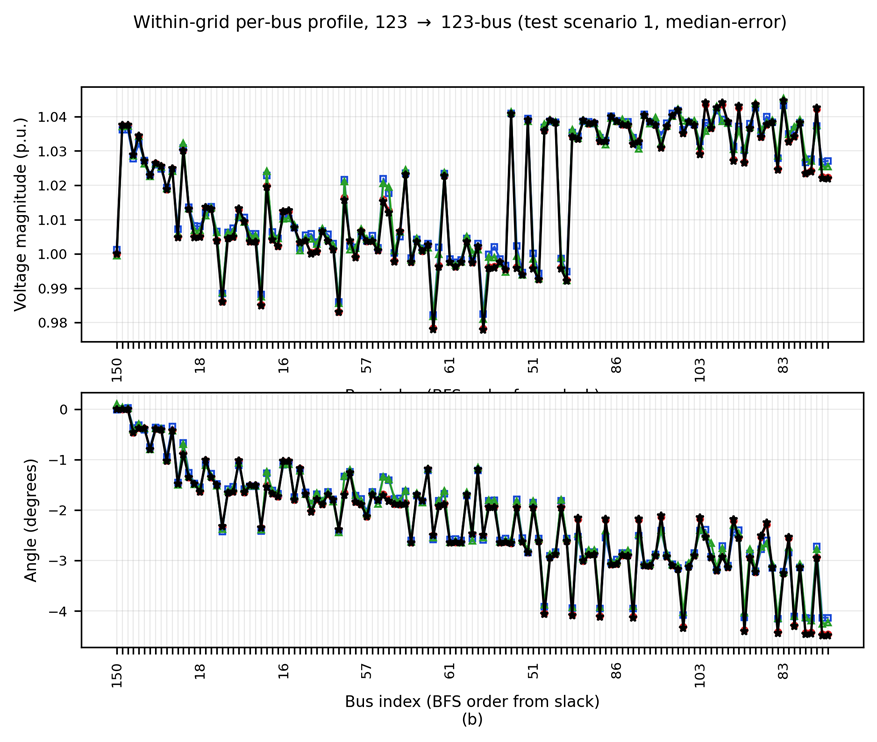}\\[-1pt]\footnotesize (c) 123-bus
\end{minipage}\hfill
\begin{minipage}[t]{0.49\textwidth}\centering
\includegraphics[width=\linewidth]{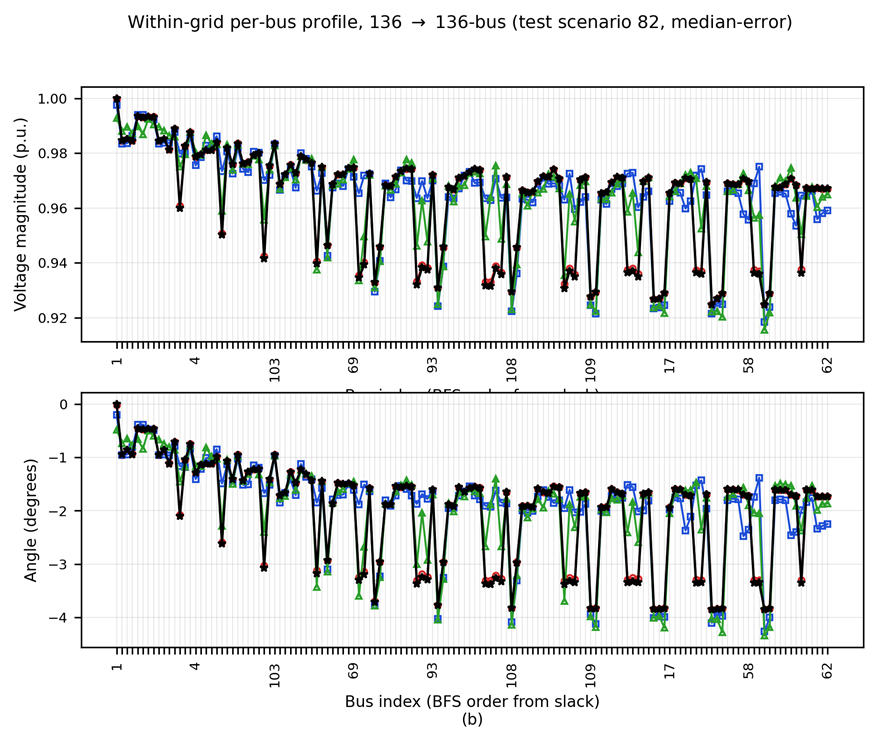}\\[-1pt]\footnotesize (d) 136-bus
\end{minipage}

\vspace{4pt}
\begin{minipage}[t]{0.49\textwidth}\centering
\includegraphics[width=\linewidth]{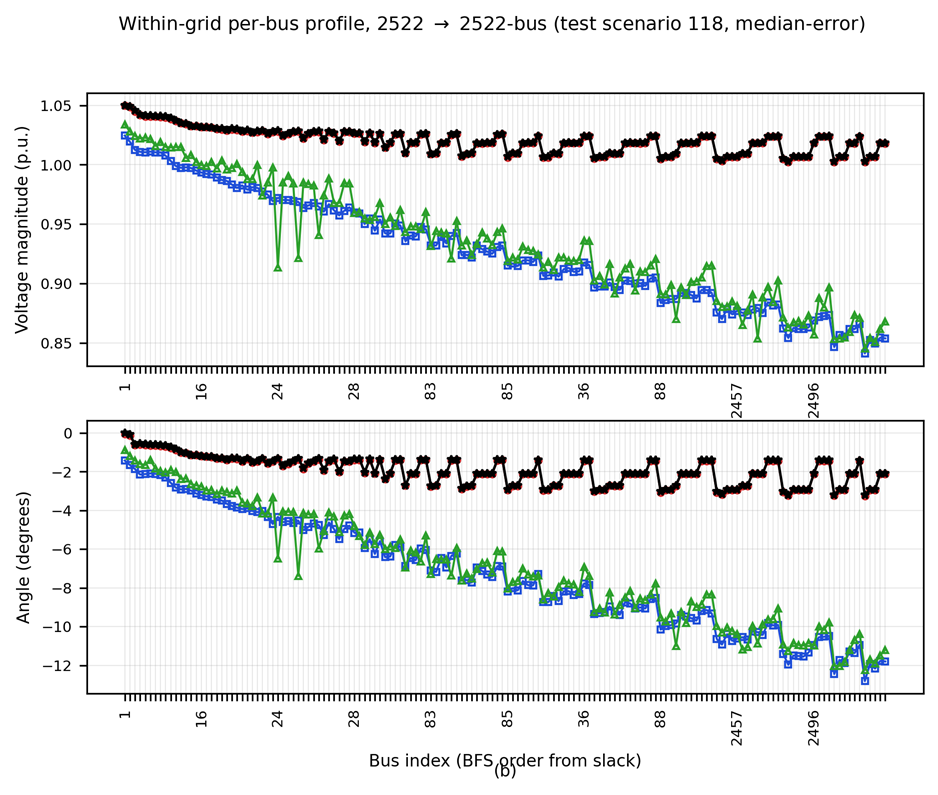}\\[-1pt]\footnotesize (e) 2522-bus
\end{minipage}
\caption{Within-grid per-bus profiles for one median-error test scenario per grid. In each panel the upper plot is voltage magnitude and the lower plot is voltage angle. Solid black lines with stars ($\star$) represent the OpenDSS ground truth. Predictions are shown for IP-GNO (red circles, $\circ$), VanillaGNO (blue squares, $\square$), and GraphSAGE (green triangles, $\triangle$). Buses are in BFS order from the slack; only every few bus labels are shown.}
\label{fig:withinprofiles}
\end{figure*}

Fig.~\ref{fig:withinprofiles} shows per-bus voltage and angle profiles of IP-GNO, VanillaGNO, and GraphSAGE in the within-grid setting, on one median-error test scenario per grid (buses in BFS order from the slack; GAT and GINE are not plotted). They give the same-grid reference for the zero-shot profiles of Section~\ref{sec:profiles} and, like those, are single scenarios that illustrate rather than replace the five-seed results of Table~\ref{tab:withingrid}.

On 33-bus (Fig.~\ref{fig:withinprofiles}a), IP-GNO is visually indistinguishable from the ground-truth voltage and angle profile, while VanillaGNO and GraphSAGE drift away from it over the last ten or so buses of the feeder (angle errors of up to roughly $0.35^{\circ}$ at the end of the feeder). On 123-bus (Fig.~\ref{fig:withinprofiles}c) all three models closely follow the ground truth, in line with the narrow margin in Table~\ref{tab:withingrid}. On 69- and 136-bus (Figs.~\ref{fig:withinprofiles}b and~\ref{fig:withinprofiles}d), IP-GNO again overlaps the ground truth over essentially the whole feeder, including the sharp bus-to-bus dips of the voltage profile, whereas VanillaGNO and GraphSAGE deviate on the far laterals, by up to roughly 0.07~pu and $1.5^{\circ}$ on 69-bus. On 2522-bus (Fig.~\ref{fig:withinprofiles}e), IP-GNO coincides with the ground truth over the whole feeder, including the repeated plateau structure of the profile. VanillaGNO and GraphSAGE instead track the overall shape but drift progressively away from it with distance from the slack: their voltage falls to roughly 0.85~pu at the far end against a true value of about 1.02~pu, and their angle error grows to roughly $8^{\circ}$--$10^{\circ}$, in line with their much larger MAE in Table~\ref{tab:withingrid}. GraphSAGE additionally shows isolated spikes of several hundredths of a per-unit.

\subsection{Zero-Shot Cross-Grid Results}
\label{sec:crossgrid}

Table~\ref{tab:crossgrid} reports all five architectures, trained once on IEEE 33-bus and evaluated zero-shot on all four unseen feeders, as mean $\pm$ standard deviation over five seeds.

\begin{table*}[t]
\centering
\caption{Zero-shot $|V|$ MAE (pu) and $\theta$ MAE (deg), trained on IEEE 33-bus only, mean $\pm$ standard deviation over 5 seeds. Bold = best per grid; (div.) = diverged out of bounds (physically implausible output, $|V|$ MAE $>0.2$~pu or $\theta$ MAE $>10^\circ$).}
\label{tab:crossgrid}
\resizebox{\textwidth}{!}{
\begin{tabular}{l cc cc cc cc cc}
\toprule
\multirow{2}{*}{Target} & \multicolumn{2}{c}{IP-GNO (Ours)} & \multicolumn{2}{c}{VanillaGNO} & \multicolumn{2}{c}{GAT} & \multicolumn{2}{c}{GINE} & \multicolumn{2}{c}{GraphSAGE} \\
\cmidrule(lr){2-3} \cmidrule(lr){4-5} \cmidrule(lr){6-7} \cmidrule(lr){8-9} \cmidrule(lr){10-11}
& $|V|$ ($\times 10^{-2}$) & $\theta$ (deg) & $|V|$ ($\times 10^{-2}$) & $\theta$ (deg) & $|V|$ ($\times 10^{-2}$) & $\theta$ (deg) & $|V|$ ($\times 10^{-2}$) & $\theta$ (deg) & $|V|$ ($\times 10^{-2}$) & $\theta$ (deg) \\
\midrule
69-bus & $\mathbf{2.32 \pm 0.28}$ & $\mathbf{0.42 \pm 0.05}$ & $7.38 \pm 0.95$ & $1.98 \pm 0.25$ & $33.7 \pm 14.5$ (div.) & $5.34 \pm 1.85$ & $86.8 \pm 21.5$ (div.) & $12.8 \pm 3.15$ & $5.13 \pm 0.65$ & $1.37 \pm 0.15$ \\
123-bus & $\mathbf{2.16 \pm 0.22}$ & $\mathbf{0.85 \pm 0.08}$ & $4.49 \pm 0.55$ & $1.48 \pm 0.12$ & $11.9 \pm 8.50$ & $1.92 \pm 0.35$ & $17.6 \pm 9.20$ & $1.97 \pm 0.42$ & $7.41 \pm 0.78$ & $1.16 \pm 0.18$ \\
136-bus & $\mathbf{3.82 \pm 0.41}$ & $\mathbf{0.88 \pm 0.09}$ & $5.70 \pm 0.65$ & $1.17 \pm 0.15$ & $14.7 \pm 12.5$ (div.) & $1.86 \pm 0.55$ & $35.3 \pm 20.5$ (div.) & $4.64 \pm 1.25$ & $4.07 \pm 0.52$ & $1.31 \pm 0.14$ \\
2522-bus & $\mathbf{3.07 \pm 0.35}$ & $\mathbf{2.95 \pm 0.25}$ & $6.51 \pm 0.85$ & $5.67 \pm 0.65$ & $6.31 \pm 1.15$ (div.) & $25.1 \pm 5.15$ & $324.0 \pm 185.0$ (div.) & $63.5 \pm 15.2$ & $1370 \pm 4550$ (div.) & $5.01 \pm 0.85$ \\
\bottomrule
\end{tabular}
}
\end{table*}

IP-GNO is the most accurate architecture on all four zero-shot target grids in both $|V|$ and $\theta$ MAE, beating the next-best non-diverged baseline in $|V|$ MAE by 54.8\% on 69-bus (vs.\ GraphSAGE), 51.9\% on 123-bus (vs.\ VanillaGNO), 6.1\% on 136-bus (vs.\ GraphSAGE), and 52.8\% on 2522-bus (vs.\ VanillaGNO). Its $\theta$ MAE ranges from $0.42^{\circ}$ to $2.95^{\circ}$ across the four targets. Only IP-GNO and VanillaGNO remain bounded on all four targets. GAT diverges on 69-, 136-, and 2522-bus, GINE on 69-, 136-, and 2522-bus, and GraphSAGE on 2522-bus ($13.7$~pu). The 136-bus margin is markedly narrower than the other three (Section~\ref{sec:discussion}). We regard IP-GNO's consistent lead, together with the instability of attention-, edge-conditioned, and neighborhood-sampling message passing under this transfer, as the paper's central empirical result.

\subsection{Per-Bus Zero-Shot Profiles}
\label{sec:profiles}

\begin{figure*}[!tbp]
\centering
\begin{minipage}[t]{0.49\textwidth}\centering
\includegraphics[width=\linewidth]{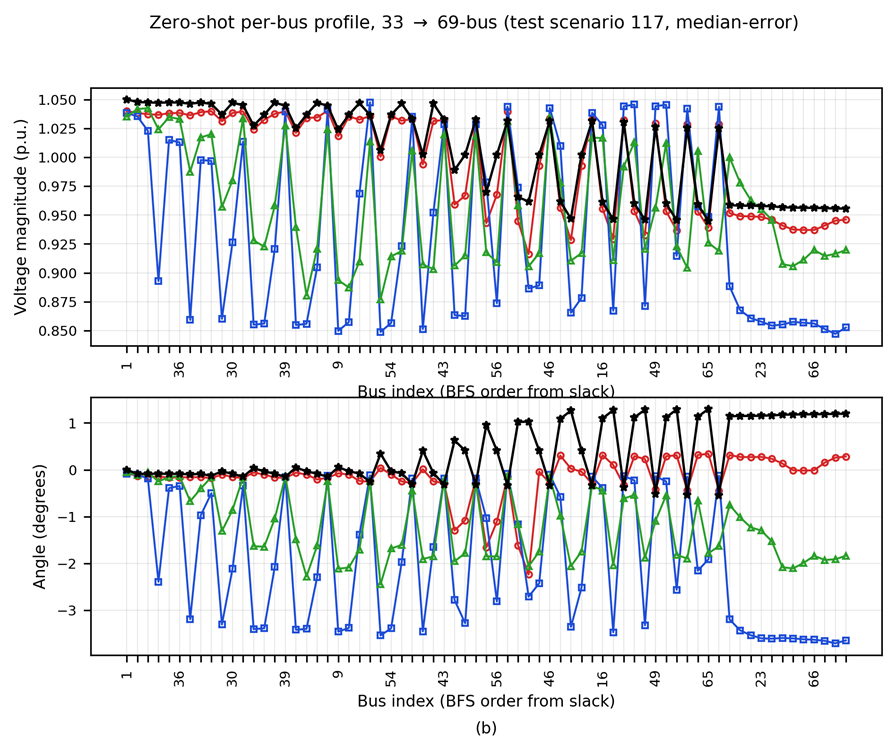}\\[-1pt]\footnotesize (a) 69-bus
\end{minipage}\hfill
\begin{minipage}[t]{0.49\textwidth}\centering
\includegraphics[width=\linewidth]{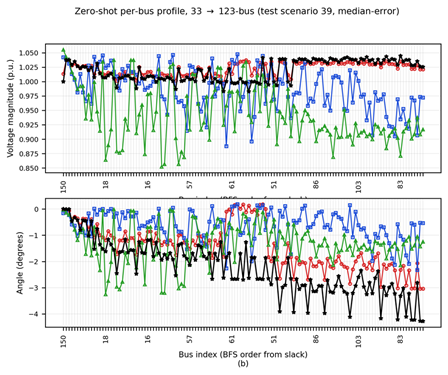}\\[-1pt]\footnotesize (b) 123-bus
\end{minipage}

\vspace{4pt}
\begin{minipage}[t]{0.49\textwidth}\centering
\includegraphics[width=\linewidth]{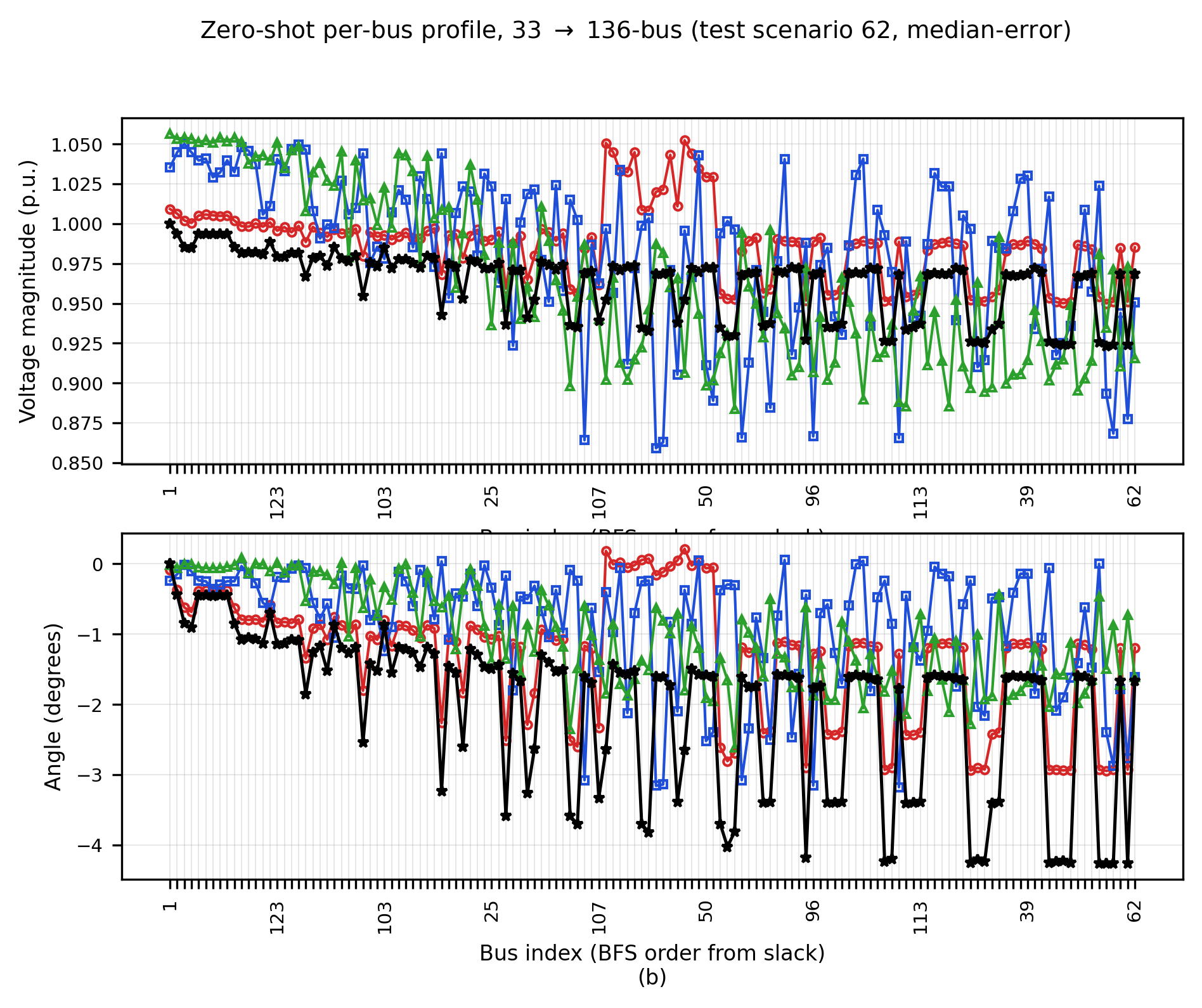}\\[-1pt]\footnotesize (c) 136-bus
\end{minipage}\hfill
\begin{minipage}[t]{0.49\textwidth}\centering
\includegraphics[width=\linewidth]{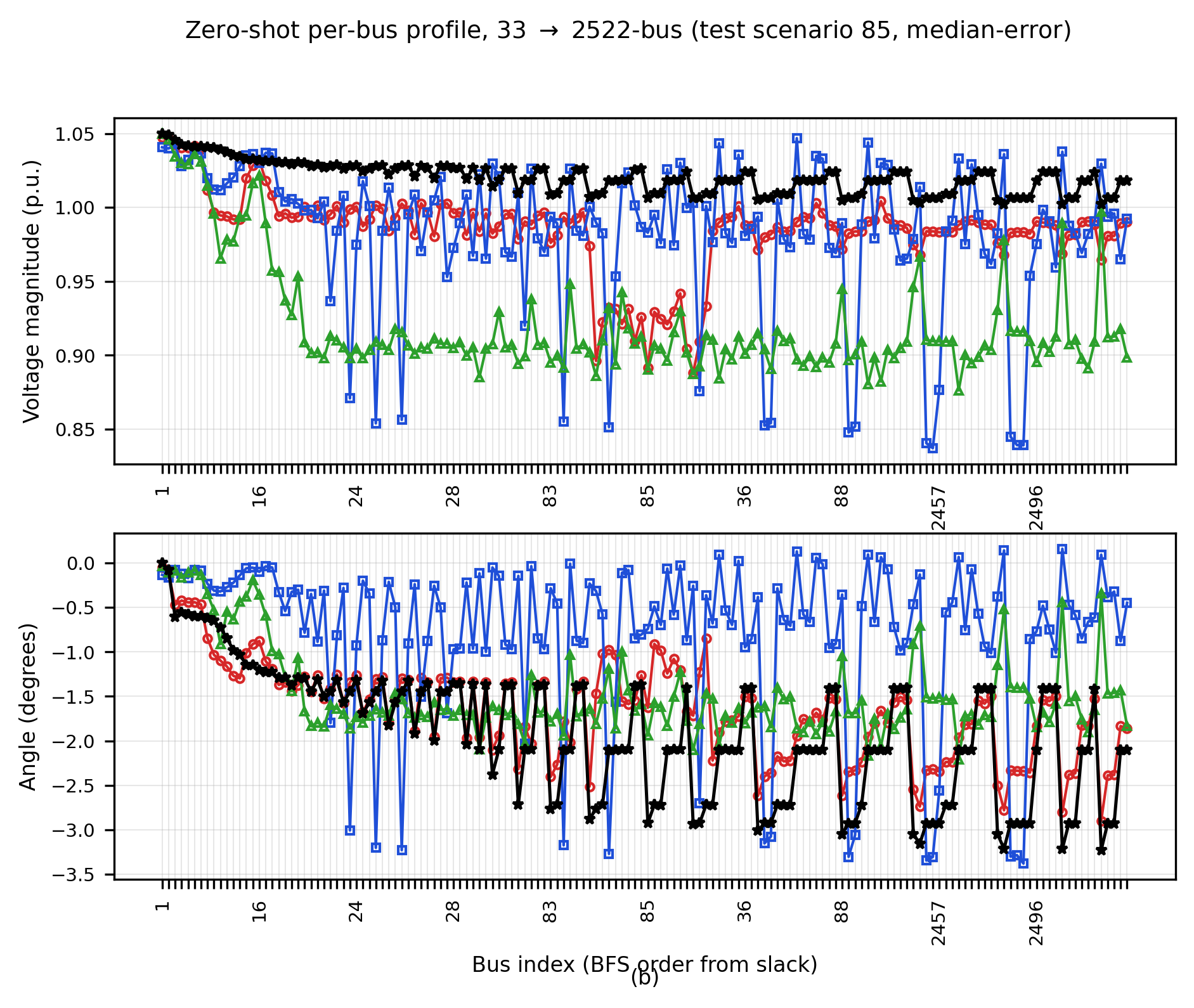}\\[-1pt]\footnotesize (d) 2522-bus
\end{minipage}
\caption{Zero-shot per-bus profiles, trained on 33-bus only, for one median-error test scenario per target grid. In each panel the upper plot is voltage magnitude and the lower plot is voltage angle. Solid black lines with stars ($\star$) represent the OpenDSS ground truth. Predictions are shown for IP-GNO (red circles, $\circ$), VanillaGNO (blue squares, $\square$), and GraphSAGE (green triangles, $\triangle$). Buses are in BFS order from the slack; only every few bus labels are shown.}
\label{fig:profiles}
\end{figure*}

Fig.~\ref{fig:profiles} shows per-bus zero-shot voltage and angle profiles of IP-GNO, VanillaGNO, and GraphSAGE on one median-error test scenario per target grid (buses in BFS order from the slack; GAT and GINE are not plotted). These are single scenarios and illustrate, rather than replace, the five-seed means of Table~\ref{tab:crossgrid}.

On 69-bus (Fig.~\ref{fig:profiles}a), IP-GNO follows both the level and the bus-to-bus oscillation of the true voltage profile along the feeder; its largest voltage deviations are confined to a few buses, where it is off by roughly 0.02--0.04~pu. VanillaGNO produces large, isolated excursions (up to roughly 0.18~pu), and GraphSAGE errs by up to about 0.15~pu on many mid-feeder buses and is biased low on the last buses. For the angle, IP-GNO reproduces the trend but underestimates the positive angles at the far end of the feeder (true $\approx+1.2^{\circ}$, predicted $\approx0^{\circ}$ to $+0.3^{\circ}$), an error of about $1^{\circ}$; the baselines' angle errors are larger on most of those buses.

On 123-bus (Fig.~\ref{fig:profiles}b), IP-GNO stays close to the true voltage profile over most of the feeder, with errors that are typically below about 0.02~pu, while VanillaGNO shows deviations of up to roughly 0.1--0.12~pu and GraphSAGE is biased low by roughly 0.1--0.2~pu over much of the feeder. IP-GNO's angle error grows toward the downstream buses, where it underestimates the (more negative) true angles by roughly $1^{\circ}$--$2^{\circ}$.

On 136-bus (Fig.~\ref{fig:profiles}c), which has the narrowest zero-shot margin (Table~\ref{tab:crossgrid}), all models are less accurate. IP-GNO captures the overall voltage level but is biased high, by roughly 0.02--0.08~pu, over the middle and downstream part of the feeder, whereas VanillaGNO and GraphSAGE scatter widely around the true profile. The mean zero-shot $\theta$ MAE of Table~\ref{tab:crossgrid} is $0.42^{\circ}$--$0.88^{\circ}$ on these three grids, roughly an order of magnitude above the within-grid values of Table~\ref{tab:withingrid} ($0.039^{\circ}$--$0.058^{\circ}$); the larger errors visible here (up to a few degrees) are concentrated on the far buses.

On 2522-bus (Fig.~\ref{fig:profiles}d), IP-GNO follows the slow decay of the true voltage and the plateau structure of the profile, but sits roughly 0.03--0.04~pu below it over most of the feeder. VanillaGNO has a comparable mean voltage deviation on this scenario but is erratic, with isolated dips to about 0.84~pu, and GraphSAGE is biased low by roughly 0.1~pu. For the angle, IP-GNO follows the trend and the large-scale oscillation of the true profile with typical errors of a few tenths of a degree, whereas VanillaGNO scatters around $-0.5^{\circ}$ and has errors of $1^{\circ}$ or more on most buses.

\subsection{Deterioration Relative to a Same-Grid Baseline}
\label{sec:deterioration}

Table~\ref{tab:deterioration} compares each target grid's zero-shot $|V|$ MAE (Table~\ref{tab:crossgrid}) against IP-GNO's \emph{own} within-grid $|V|$ MAE on that grid (Table~\ref{tab:withingrid}).

\begin{table}[t]
\centering
\caption{Zero-shot deterioration factor of IP-GNO, relative to its own within-grid performance on each target feeder}
\label{tab:deterioration}
\begin{tabular}{lccc}
\toprule
Grid & Within-grid & Zero-shot & Factor \\
\midrule
69-bus & $2.67\times10^{-3}$ & $2.32\times10^{-2}$ & $8.70\times$ worse \\
123-bus & $1.88\times10^{-3}$ & $2.16\times10^{-2}$ & $11.5\times$ worse \\
136-bus & $9.29\times10^{-4}$ & $3.82\times10^{-2}$ & $41.1\times$ worse \\
2522-bus & $8.40\times10^{-4}$ & $3.07\times10^{-2}$ & $36.5\times$ worse \\
\bottomrule
\end{tabular}
\end{table}

Across the four target grids, zero-shot error is 8.7--41.1$\times$ IP-GNO's own within-grid error, a substantial generalization gap that varies by feeder.

\subsection{Ablation Study}
\label{sec:ablation}

Ablation confirms the DC angle prior and source-injection skip are the most load-bearing components of the architecture. On the IEEE 33-bus full-graph training protocol, removing the DC angle prior raises $|V|$ MAE by 292.5\% and $\theta$ MAE by 135.8\%. Removing the source-injection skip raises $|V|$ MAE by 270.4\% and $\theta$ MAE by 113.5\%. Other PETE positional features contribute marginally individually, and an explicit self-admittance gate at every kernel layer was found to be redundant given the zeroth-order gate.

\subsection{Gaussian Input-Measurement-Noise Robustness}
\label{sec:noise}

\textbf{Noise setup.} To emulate SCADA/AMI measurement error, Gaussian noise is added to the input load measurements, with the standard deviation set as a fraction $\sigma$ of each bus's own $P$ or $Q$ magnitude:
\begin{equation}
\begin{split}
\tilde P_i &= P_i + \sigma\,|P_i|\,\epsilon_i, \qquad \tilde Q_i = Q_i + \sigma\,|Q_i|\,\epsilon_i',\\
&\epsilon_i,\epsilon_i' \sim \mathcal{N}(0,1),
\end{split}
\end{equation}
for $\sigma \in \{1,5,10,20,30\}\%$. The noise is applied exclusively to the input load measurements fed to the network, while the ground-truth $|V|$ and $\theta$ targets remain clean. The noise is an evaluation-only perturbation applied at test time to a single trained checkpoint (seed~0); no model is trained or fine-tuned on noisy inputs, and no denoising component is involved.

Table~\ref{tab:noise} reports the percentage deterioration of within-grid $|V|$ MAE relative to the noise-free case, where the 0\% column is the noise-free reference (100\%).

\begin{table}[t]
\centering
\caption{Within-Grid Percentage Deterioration of $|V|$ MAE under Input Noise}
\label{tab:noise}
\footnotesize
\setlength{\tabcolsep}{3pt}
\resizebox{\columnwidth}{!}{%
\begin{tabular}{lcccccc}
\toprule
Grid & 0\% & 1\% & 5\% & 10\% & 20\% & 30\% \\
\midrule
69-bus & 100.0\% & +0.0\% & +0.1\% & +0.2\% & +0.5\% & +0.9\% \\
123-bus & 100.0\% & +0.0\% & +0.0\% & +0.1\% & +0.2\% & +0.3\% \\
136-bus & 100.0\% & +0.0\% & +0.0\% & +0.0\% & +0.1\% & +0.2\% \\
2522-bus & 100.0\% & +0.0\% & +0.0\% & +0.1\% & +0.2\% & +0.4\% \\
\bottomrule
\end{tabular}}
\end{table}

Deterioration stays below 1\% on every grid even at 30\% noise (at most $+0.9\%$ on 69-bus and $+0.4\%$ or less on the other three). A flat response could also indicate that the prediction is only weakly sensitive to the load inputs over the sampled range; this test does not separate the two readings, and a direct sensitivity analysis of the predicted voltages with respect to the injections (for example against $Z_{\bus}$) is a natural next step.

\section{Discussion}
\label{sec:discussion}

\textbf{Headline result.} IP-GNO is the most accurate architecture on every zero-shot target grid and on all five within-grid settings, and together with VanillaGNO it is the only architecture that stays bounded on all four zero-shot targets. The divergence of GAT (69-, 136-, and 2522-bus), GINE (69-, 136-, and 2522-bus), and GraphSAGE (2522-bus) shows that these message-passing schemes can become numerically unreliable under cross-grid transfer, whereas a kernel parameterized directly by admittance (IP-GNO), or one with a clean identity path (VanillaGNO), remains bounded. The study covers balanced, positive-sequence radial feeders and a single training feeder (33-bus); other training feeders, unbalanced multi-phase models, and meshed networks are outside its scope.

\textbf{Narrower margins.} IP-GNO's zero-shot margin is 51--55\% on three grids but 6.1\% on 136-bus, and its within-grid margin is 2.1\% on 123-bus. The 136-bus feeder has the largest load in the study and materially different impedance heterogeneity across its laterals; per-graph normalization removes admittance \emph{scale} but not a shift in relative heterogeneity, and GraphSAGE, which does not use edge features, is comparatively insulated from such a shift, which may explain why it is the closest competitor. On 123-bus, every model is fit to one fixed topology, which may leave less room for an admittance-aware kernel to help. Both explanations are hypotheses motivated by the structural differences between the grids, not tested mechanisms.

\textbf{Scope, limitations, and future work.} Four points bound how the results should be read. \emph{(i) Accuracy regime.} Zero-shot $|V|$ MAE is $2.2\times10^{-2}$--$3.8\times10^{-2}$~pu, 8.7--41.1$\times$ IP-GNO's own within-grid error (Table~\ref{tab:deterioration}); the raw output is therefore best viewed as a screening-grade estimate or an initial guess. \emph{(ii) Evidence base.} Results are means $\pm$ standard deviations over five seeds; we did not run formal significance tests, so margins of a few percent (123-bus within-grid, 2.1\%; 136-bus zero-shot, 6.1\%) should be read as ``comparable'' rather than decisive. The explanations offered for the 136-bus and 123-bus behaviour are hypotheses. \emph{(iii) Comparisons not made.} We compare against four message-passing baselines under a matched protocol; we do not compare against training-free linearized power-flow models (e.g., LinDistFlow) or other learned power-flow solvers, nor do we isolate the admittance-feature input from the gate and skip in a factorial ablation. \emph{(iv) Model scope.} Training uses a single feeder; all networks are balanced, positive-sequence, and radial, and the load scenarios contain no distributed generation or reverse power flow. Multi-feeder training, unbalanced three-phase models, DER-rich operating points, and a sensitivity analysis against $Z_{\bus}$ are the natural extensions.

\section{Conclusion}

The AC power-flow solution operator possesses an algebraic structure -- the Neumann expansion of $Y_{\bus}^{-1}$ -- that is independent of graph size, and a graph neural operator explicitly aligned with this structure transfers across feeders of different size far better than architecture-agnostic GNNs. IP-GNO, trained on full graphs of the IEEE 33-bus feeder, is the most accurate of five architectures on all four structurally distinct, unseen zero-shot target feeders (69-, 123-, 136-, and 2522-bus), and together with VanillaGNO is the only one that remains bounded and physically plausible on all four; GAT and GINE diverge on several targets and GraphSAGE diverges on 2522-bus. Within-grid, IP-GNO is the most accurate on all five feeders (33-, 69-, 123-, 136-, and 2522-bus). The ablation identifies the DC angle prior and the source-injection skip as the two most load-bearing components (removal raises $|V|$ MAE by $292.5\%$ and $270.4\%$), and shows that an explicit self-admittance gate at every kernel layer is redundant given the zeroth-order gate, most plausibly because the residual stream already carries that information. Under Gaussian input-measurement noise of up to 30\%, within-grid $|V|$ MAE deteriorates by less than 1\%. The study is limited to balanced radial feeders and a single training feeder (Section~\ref{sec:discussion}); extending the training set, the network model, and the comparison set is the main direction for future work.

\bibliographystyle{IEEEtran}
\bibliography{references}

\end{document}